\documentclass[a4paper,fleqn]{cas-sc}

\usepackage[authoryear,longnamesfirst]{natbib}
\usepackage[utf8]{inputenc}
\usepackage{subcaption}
\usepackage{float}
\usepackage[section]{placeins} 

\begin{document}
\let\WriteBookmarks\relax

\shortauthors{P. Gupta and P.K. Gupta}

\shorttitle{Deep learning architectures for brain tumor classification}

\title[mode=title]{Comparative Study of Deep Learning Architectures for Brain Tumor Classification}

\author[1]{Prisha Gupta}
\cormark[1]

\affiliation[1]{organization={Amrita School of Artificial Intelligence, Amrita Vishwa Vidyapeetham},
    city={Coimbatore},
    country={India}}

\author[2]{Pradeep Kumar Gupta}

\affiliation[2]{organization={Department of Computer Science, Jaypee University of Information Technology},
    city={Solan},
    state={Himachal Pradesh},
    country={India}}

\cortext[1]{Corresponding author}

\begin{abstract}
Accurate and early classification of brain tumors from magnetic resonance imaging (MRI) is essential for treatment planning, yet many published deep learning studies are trained and tested on small, single-source datasets and report accuracies that may not hold outside those datasets. This study compares six deep learning architectures for four-class classification (healthy, meningioma, glioma and pituitary tumor): a baseline convolutional neural network (CNN), an attention-based CNN, ShuffleNet, and EfficientNet-B0, ResNet-50 and Xception augmented with the Convolutional Block Attention Module (CBAM). The models were trained on a corpus assembled from four publicly available MRI datasets. Because these datasets carry no patient identifiers, a leakage-aware pipeline was developed that removes exact duplicates using MD5 hashing (550 images) and near-duplicates using perceptual hashing (8,024 images), and assigns visually similar images to the same fold during 5-fold group cross-validation. A leave-one-dataset-out (LODO) test on a fully held-out dataset was used to measure cross-dataset generalization. Under a conventional random split, ShuffleNet achieved 97.1\% accuracy with 1.2M parameters. Under leakage-aware group K-fold cross-validation, ShuffleNetV2 ($98.41 \pm 0.16$\%) and EfficientNet-B0 ($98.23 \pm 0.21$\%) were statistically indistinguishable (paired t-test, $p = 0.19$), and both significantly outperformed the custom CNNs. On the held-out dataset, accuracy dropped for every model: EfficientNet-B0 generalized best (78.4\%), whereas ShuffleNetV2 fell to 66.2\%. Glioma--meningioma confusion was the dominant error mode, and Grad-CAM maps showed that predictions were driven by tumor regions. These results show that in-distribution accuracy alone overstates real-world performance and that leakage-aware and cross-dataset evaluation should be reported alongside it.
\end{abstract}



\maketitle

\section{Introduction}

Brain tumors are among the most life-threatening neurological disorders because of their aggressive growth and the complexity of their presentation. Early and accurate diagnosis is important for timely treatment and improves patient survival rates. Magnetic resonance imaging (MRI) is considered the gold standard for the non-invasive detection and classification of brain tumors, as it provides detailed visualization of tumor morphology in a short acquisition time.

However, manual interpretation of MRI scans is time-consuming and prone to error, particularly for early-stage lesions and for tumor types that differ only subtly, such as meningiomas and gliomas. This has highlighted the need for automated tools that support radiologists in decision-making. The advent of machine learning and deep learning, particularly convolutional neural networks (CNNs), has transformed medical image analysis and enabled high accuracy in brain tumor identification and classification from imaging data.

\subsection{CNN-Based Approaches for Brain Tumor Classification}
\textbf{Traditional CNN architectures:}
Modern brain tumor classification (BTC) has been built largely on CNN-based models with various architectural modifications \citep{sultan2019multi,kibriya2022multiclass,Garg_2023,Singh_2023}, which have shown high success rates in tumor detection and classification.
\citet{sultan2019multi} were among the first to apply CNNs to BTC, working with two datasets: in the first, tumors were classified into three primary types (meningioma, glioma and pituitary tumor), and in the second, into three glioma grades. Their study achieved overall accuracies of 96.13\% and 98.7\% on the first and second datasets, respectively.

Building on this foundation, \citet{kibriya2022novel} proposed a 13-layer CNN designed specifically for this application and achieved 97.2\% accuracy, suggesting that deeper architectures capture complex tumor features more effectively. \citet{Garg_2023} developed a 14-layer 2D deep CNN that achieved a test accuracy of 98.66\%, showing continued improvement through architectural refinement.

\citet{badvza2020classification} worked with T1-weighted contrast-enhanced MRI images, as did several later studies \citep{alhassan2021brain,tummala2022classification,mzoughi2025vision}. They performed record-wise and subject-wise 10-fold cross-validation on both the original and augmented image databases, and found that the record-wise approach gave a higher accuracy (96.56\%) than the subject-wise approach.

\textbf{Multi-task CNN models:}
Extending beyond single-task classification, \citet{Singh_2023} used three distinct CNN models for separate categorization tasks. The first model classified brain tumors with an accuracy of 99.74\%. The second achieved 96.27\% accuracy in differentiating normal brain tissue from glioma, meningioma, pituitary and metastatic tumors. The third distinguished tumor grades II, III and IV with an accuracy of 99.18\%, highlighting the versatility of CNNs across classification scenarios.
\citet{gull2023deep} performed a similar study on a combination of three datasets (Kaggle, Figshare and REMBRANDT). Their first model separated tumor from non-tumor images with 99.40\% accuracy, the second performed three-class classification of meningioma, glioma and pituitary tumors with 97.78\% accuracy, and the third classified four tumor grades (I--IV) with 98.91\% accuracy. \citet{alhassan2021brain} proposed a hard-swish-based ReLU activation function within a CNN, achieving 98.6\% accuracy in BTC.

\subsection{Advanced Deep Learning Architectures}
\textbf{Densely connected networks:}
\citet{prakash2023classification} introduced a densely connected convolutional network (DenseNet) for BTC, evaluated on a public dataset of 3,064 MR images covering meningioma, glioma and pituitary tumors \citep{kibriya2022multiclass}. DenseNet outperformed VGG16, SVM and AlexNet, achieving an accuracy of 97.39\%.

\textbf{Hierarchical deep learning approaches:}
\citet{shajin2023efficient} proposed a framework for the BraTS dataset \citep{khan2023automated} that used a hierarchical deep learning neural network (HieDNN) classifier to handle complex medical imaging data and perform BTC.

\subsection{Transfer Learning and Pre-trained Models}
\textbf{EfficientNet-based transfer learning:}
Transfer learning has shown considerable success in brain tumor classification. \citet{Zulfiqar_2023} fine-tuned pre-trained EfficientNets and achieved an overall test accuracy, precision, recall and F1-score of 98.86\%, 98.65\%, 98.77\% and 98.71\%, respectively, on the Figshare brain tumor dataset, a dataset also used by \citet{Mahum_2023}, \citet{babu2023detection} and \citet{gull2023deep}. \citet{babu2023detection} subsequently fine-tuned five EfficientNet variants (B0--B4) and found that EfficientNetB2 gave the best results, with an accuracy of 99.06\%, precision of 98.73\%, recall of 99.13\% and F1-score of 98.79\%. \citet{owida2024automated} derived a new architecture from EfficientNetB3 that reached 100\% accuracy on a dataset of 7,023 MRI scans.

\textbf{Enhanced deep transfer learning methods:}
\citet{masood2022brain} combined attention with an established architecture, using ResNet34 enhanced with attention blocks as the base network for brain tumor identification and classification. Their approach achieved 98.98\% accuracy, demonstrating the value of attention in focusing on relevant image regions.
\citet{asif2023enhanced} conducted two experiments with five models: Xception, DenseNet201, DenseNet121, ResNet152V2 and InceptionResNetV2. The first experiment classified three classes (meningioma, glioma and pituitary), and the second added a fourth class for healthy subjects. The Xception-based model achieved the highest accuracy in both: 99.67\% on the three-class task and 95.87\% on the four-class task.
\citet{malkauthekar2025novel} presented a Probabilistic Bi-Level Teaching--Learning-Based Optimization (P-BTLBO) algorithm for hybrid feature extraction and multi-class brain tumor classification using ResNet-50 and GLCM features.
\citet{kanna2025new} proposed a penguin-search-optimized quantum-enhanced support vector machine (PSO-QESVM) that used ResNet and VGG16 to extract features from the pre-processed data, achieving an accuracy of 98.90\%, recall of 98.9\%, F1-score of 98.5\% and precision of 98.7\%.

\subsection{Hybrid Approaches and Feature Fusion}
\textbf{Segmentation--classification integration:}
\citet{Mahum_2023} combined segmentation and feature fusion for brain tumor detection. Working with the Figshare and Harvard datasets \citep{Mostafa_2023,tummala2022classification}, they used a bidirectional LSTM network to classify tumors into pituitary, glioma and meningioma.
Because segmentation is time-consuming, \citet{Mostafa_2023} instead pre-processed the images with a Gaussian filter (GF) and SynthStrip to improve localization, achieving 99.8\% accuracy, 99.3\% recall, 99.4\% precision, a 99.5\% F1-score and an AUC of 0.989.
\citet{khan2023automated} presented a hybrid framework that used a U-Net for segmentation and an SVM to separate normal from tumor-affected images, achieving an overall accuracy of 98.95\%, specificity of 98.13\% and sensitivity of 97.46\%.
\citet{ozyurt2020expert} used fuzzy C-means with super-resolution (SR-FCM) to detect tumors in low-resolution MRI scans, extracted features with SqueezeNet, and classified them with an extreme learning machine (ELM), achieving an overall accuracy of 98.33\%.

\textbf{Ensemble methods and deep feature integration:}
\citet{kang2021mri} used 13 pre-trained deep CNNs as feature extractors together with nine machine learning (ML) classifiers. The three best-performing deep features were concatenated into an ensemble feature set that was then fed back into the ML classifiers.
\citet{Sandhiya_2023} performed a similar study using deep CNN (DCNN) features and ML classifiers, comparing ten pre-trained DCNN feature extractors with five ML classifiers. An SVM with an RBF kernel on DenseNet features gave the best performance.
\citet{alturkicombining} extracted features with a deep CNN and classified them with a voting classifier that combined logistic regression and stochastic gradient descent, achieving an accuracy of 99.9\%.

\textbf{Vision transformer ensembles:}
\citet{tummala2022classification} used an ensemble of Vision Transformer (ViT) models for three-class classification. The best individual model (L/32) achieved 98.2\% accuracy, whereas the ensemble of all four ViT models achieved 98.7\%.
\citet{tanone2025vit} proposed ViT-CB, which used a ViT for feature extraction, PCA for dimensionality reduction and CatBoost for classification, with SHapley Additive exPlanations (SHAP) used to interpret the results. The method achieved accuracies of 99.32\% and 90.00\% on its two evaluation datasets.
\citet{chandraprabha2025novel} presented an end-to-end ViT--CNN architecture that achieved 99.64\% accuracy on a dataset of 4,855 images spanning several tumor classes.
\citet{mzoughi2025vision} trained a deep ViT for multi-class classification of common primary tumors (gliomas, meningiomas and pituitary tumors) and explained its predictions with Gradient-weighted Class Activation Mapping (Grad-CAM), Local Interpretable Model-agnostic Explanations (LIME) and SHAP. The CNN and ViT achieved test accuracies of 83.37\% and 91.61\%, respectively.

\textbf{Deep CNN feature fusion:}
\citet{kibriya2022multiclass} presented a multiclass BTC method based on deep feature fusion, extending their previous work. Features from AlexNet, GoogLeNet and ResNet18 were fused into a single feature vector and classified with a support vector machine (SVM) and K-nearest neighbors (KNN), achieving an accuracy of 99.7\%.
\citet{Haq_2023} proposed DCNNBT, a deep CNN for brain tumor detection and classification that outperformed six established pre-trained models (SE-ResNet-101, SE-ResNet-152, SENet-154, ResNet152V2, EfficientNetB0 and EfficientNetB5) with an accuracy of 99.18\%.
\citet{albalawi2024integrated} combined CNNs with a federated learning framework for BTC, using a modified VGG16 and highlighting the value of federated and transfer learning in medical imaging. This approach achieved an accuracy of 98\%.

\subsection{Research Gaps}
The literature review identified several critical gaps. First, most studies used relatively small, single-source datasets. This limits diversity in patient population and imaging conditions and is compounded by the lack of standardized evaluation protocols, highlighting the need for larger datasets that combine multiple sources.

Second, there is a gap between the high accuracies reported in research settings and the performance expected in clinical practice, and many complex ensemble and hybrid approaches may be impractical for real-time deployment. Little attention has been paid to optimizing models for clinical environments with computational constraints and diverse patient populations.

Third, evaluation methodologies and preprocessing approaches are inconsistent across studies, and few comparisons use identical datasets and evaluation criteria, which makes results difficult to compare. Finally, although some recent works incorporate explainable AI (XAI) techniques, many earlier approaches lacked interpretability, which limits clinical validation and motivates better integration of explainability methods.

\subsection{Objectives}
The primary objective of this research is to develop a comprehensive and clinically applicable deep learning framework for automated brain tumor classification that addresses these limitations. Specifically, this study aims to:
\begin{itemize}
    \item integrate four publicly available datasets into a large, diverse brain tumor dataset with standardized preprocessing;
    \item systematically evaluate and compare deep learning and transfer learning architectures under identical evaluation protocols;
    \item identify models that balance classification accuracy with computational efficiency for real-time clinical deployment;
    \item establish leakage-aware evaluation protocols that measure generalization across imaging sources; and
    \item achieve strong multi-class classification performance while addressing the methodological inconsistencies identified in the literature.
\end{itemize}

\section{Experimental Setup}
\subsection{Hardware and Software Environment}
\subsubsection{Computing Infrastructure}
Initial experiments were conducted on a laptop with an NVIDIA GeForce GTX 1650 GPU, an 11th-generation Intel Core i5-11400H processor (2.70\,GHz), 16\,GB of RAM and 477\,GB of storage, running a 64-bit operating system. Because of these hardware limitations, the combined dataset could not be loaded into memory at once. Batch sizes were therefore limited to 8--16 images, data generators were used to load images on demand, preprocessing was performed in smaller chunks, and model checkpoints were saved frequently to prevent loss of progress during long training sessions. These measures increased the overall training time.

\subsubsection{Software Stack}
To overcome the memory limitations of the local setup, model development and training were migrated to the Kaggle platform, which provides NVIDIA Tesla P100 and T4 GPUs with substantially more memory than the local GTX 1650. The software stack comprised TensorFlow 2.8.2 with Keras 2.8.0 and PyTorch for model implementation, supported by NumPy 1.21.6 for array operations, Pandas 1.3.5 for data management, OpenCV 4.6.0 for image processing and Scikit-learn 1.0.2 for evaluation metrics. Visualization and analysis were performed with Matplotlib 3.5.3 and Seaborn 0.11.2.

\subsection{Evaluation Protocol}
\subsubsection{Cluster-Aware Group K-Fold Cross-Validation}
\label{sec:groupkfold}
The source datasets do not provide patient identifiers, so images from the same patient, or duplicated images, could appear in both training and test sets and inflate the reported performance through data leakage. To obtain a robust, leakage-free estimate of model performance, a 5-fold Group K-Fold cross-validation strategy was employed. Unlike standard K-Fold validation, Group K-Fold ensures that samples belonging to the same group are never split across training and validation folds.

Groups were defined by perceptual hash (pHash) clustering of the deduplicated images (Section~\ref{sec:clusters}), so that visually similar MRI slices were assigned to the same cluster. These cluster identifiers were used as grouping constraints, and all images in a given cluster were therefore assigned to a single fold, preventing leakage from highly correlated samples.

Formally, let $\mathcal{D} = \{(x_i, y_i, g_i)\}_{i=1}^{N}$ denote the dataset, where $x_i$ is an image, $y_i$ its class label and $g_i$ its cluster identifier. Group K-Fold partitions $\mathcal{D}$ into $K=5$ disjoint folds such that
\[
g_i \in \text{Fold}_k \implies g_i \notin \text{Fold}_{j}, \quad \forall j \neq k.
\]
In each round, models were trained on four folds and evaluated on the remaining fold. Performance metrics were averaged across the five folds.

\subsubsection{Leave-One-Dataset-Out (LODO) Evaluation}
\label{sec:lodo}
To assess cross-dataset generalization, a Leave-One-Dataset-Out (LODO) protocol was adopted, in which all images from one source dataset were held out entirely for testing while the remaining datasets were used for training. Let $\mathcal{D} = \bigcup_{m=1}^{M} \mathcal{D}_m$ denote the combined dataset constructed from $M=4$ independent sources. The model was trained on $\mathcal{D} \setminus \mathcal{D}_m$ and evaluated on the unseen source $\mathcal{D}_m$. The Lakshan dataset (Dataset~3) was used as the held-out source because it is the only one that contains all four classes; the other sources each lack at least one class and therefore cannot serve as a four-class test set.

\section{Datasets and Preprocessing}
\subsection{Dataset Construction}
\begin{figure}[!htbp]
    \centering
    \includegraphics[width=0.7\linewidth]{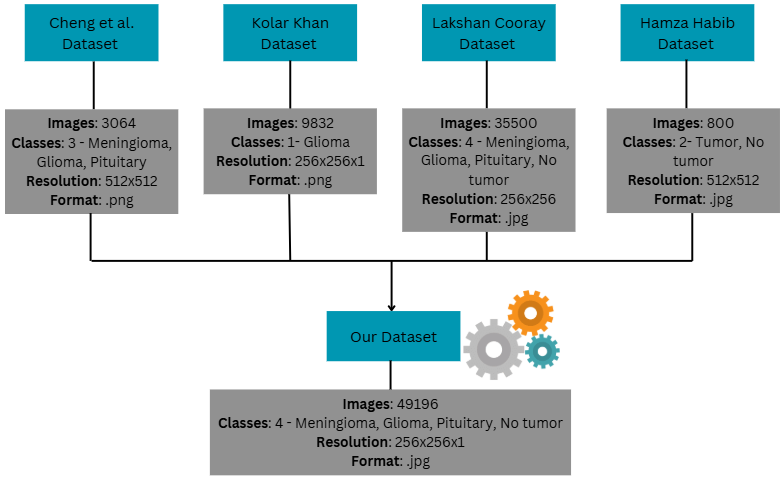}
    \caption{Construction of the combined dataset from four public sources.}
    \label{fig:dataset}
\end{figure}

This study used four publicly available brain tumor MRI datasets to build a large and diverse training corpus (Fig.~\ref{fig:dataset}), addressing the limitation of single-source datasets identified in the literature review. Merging these datasets increased both the quantity and the diversity of the data, which is expected to improve the robustness of models trained on it.

\textbf{Dataset 1:} This dataset contains 3,064 T1-weighted contrast-enhanced images from 233 patients with three kinds of brain tumor: meningioma (708 slices), glioma (1,426 slices) and pituitary tumor (930 slices), at a resolution of $512\times512$ pixels \citep{Cheng2017-fj}. The files were converted from the \texttt{.mat} format to \texttt{.png} by the Kaggle uploader.

\textbf{Dataset 2:} The dataset by \citet{kolar_khan_2023} contains 9,832 MRI brain tumor images focused on glioma classification and detection. The images are $256\times256$ grayscale in \texttt{.jpg} format.

\textbf{Dataset 3:} This dataset \citep{lakshancooray} is pre-divided into training and test folders, each containing four classes (glioma, meningioma, pituitary and no tumor), with a total of 35,500 brain MRI images. Resolution varies across images and was standardized during preprocessing.

\textbf{Dataset 4:} This dataset \citep{kaggleBrainCancer} contains 800 \texttt{.jpg} images of healthy and tumor-affected brains at variable resolution.

As shown in Fig.~\ref{fig:dataset}, the four datasets were combined into a single dataset with four classes (healthy, meningioma, glioma and pituitary tumor), which were mapped to separate directories (D0, D1, D2 and D3, respectively) for the training, validation and test splits.

\subsection{Standard Training Preprocessing Pipeline (TensorFlow)}
\label{sec:standard-pipeline}
This pipeline was used for model training under standard train--validation--test splits, reflecting common practice in brain tumor classification studies. It was implemented in TensorFlow--Keras to take advantage of efficient data loading and on-the-fly augmentation through the \texttt{tf.data} API. Parallel processing and prefetching with \texttt{tf.data.AUTOTUNE} minimized I/O bottlenecks, and a fixed random seed (42) was used for shuffling to ensure reproducibility.

\textbf{Image loading and standardization:}
Each image was preprocessed to ensure uniform size, format and value range:
\begin{itemize}
    \item images were read from disk in common formats (\texttt{.jpg}, \texttt{.jpeg}, \texttt{.png});
    \item images were decoded into three-channel RGB tensors;
    \item images were resized to $224\times224$ pixels; and
    \item pixel values were converted to 32-bit floating point and normalized to $[0, 1]$ by dividing by 255.
\end{itemize}

\textbf{Data augmentation:}
To improve generalization and robustness, data augmentation was applied to the training split only. Each of the following random transformations was applied with a probability of 50\%:
\begin{itemize}
    \item random horizontal or vertical flips, to simulate variations in orientation;
    \item brightness adjustment of up to 20\%, to account for varying imaging conditions;
    \item contrast adjustment within $[0.8, 1.2]$;
    \item saturation adjustment within $[0.8, 1.2]$;
    \item random rotation between $-15^\circ$ and $15^\circ$, to simulate misalignment;
    \item random zoom by a factor between 0.9 and 1.1, followed by cropping or padding to restore the original resolution; and
    \item additive random noise with a standard deviation of 0.05.
\end{itemize}

\subsection{Leakage-Aware Preprocessing and Evaluation Pipeline (PyTorch)}
\label{sec:leakage-pipeline}
\begin{figure}[!htbp]
    \centering
    \includegraphics[width=\linewidth]{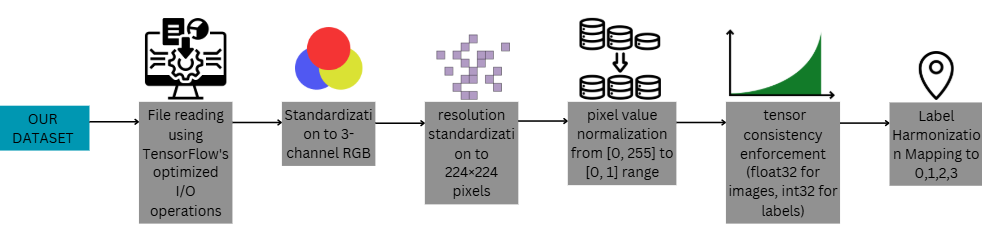}
    \caption{Leakage-aware dataset preprocessing pipeline.}
    \label{fig:preprocessing}
\end{figure}

Although the standard pipeline was sufficient for baseline evaluation, it did not address the data leakage risks that arise from aggregating multiple sources, including overlapping patients and inconsistent labeling. To enable robust cross-validation and cross-dataset analysis, an additional leakage-aware preprocessing pipeline was developed (Fig.~\ref{fig:preprocessing}).

All four datasets were first consolidated into a single metadata table and mapped to a common label space of four classes, healthy (0), meningioma (1), glioma (2) and pituitary (3), to ensure consistent supervision across sources.

\subsubsection{Exact Duplicate Removal using MD5 Hashing}
Overlap between the source datasets produces identical images. Exact duplicates were detected using MD5 hashing: images with identical hash values were treated as byte-wise duplicates, and only the first occurrence was retained so that redundant samples could not bias training or evaluation. In total, 550 exact duplicates were found and removed.

\subsubsection{Near-Duplicate Removal using Perceptual Hashing}
Beyond exact duplicates, perceptually similar images, such as resized, contrast-modified or lightly transformed versions of the same slice, were identified using perceptual hashing (pHash). Images were converted to grayscale, resized to a uniform resolution, hashed, and compared using the Hamming distance between their binary hash vectors:
\begin{equation}
    d_h(\mathbf{h}_i, \mathbf{h}_j) = \sum_{k=1}^{n} \left| h_i^{(k)} - h_j^{(k)} \right|,
    \label{eq:hamming}
\end{equation}
where $h_i^{(k)} \in \{0,1\}$ is the $k$-th bit of the $n$-bit hash of image $i$. Pairs with $d_h \le 3$ were treated as near-duplicates, and only one representative image per group was kept. This substantially reduces the risk that visually similar slices appear in both training and test splits. In total, 8,024 near-duplicates were detected and removed.

\subsubsection{Image Standardization and Enhancement}
Each retained image underwent a standardized preprocessing routine:
\begin{itemize}
    \item conversion to grayscale;
    \item resizing to $224\times224$ pixels;
    \item contrast enhancement using contrast-limited adaptive histogram equalization (CLAHE);
    \item intensity normalization using an interquartile range (IQR)-based standardization scheme, which is less sensitive to outliers than min--max scaling; and
    \item conversion to a three-channel format for compatibility with ImageNet-pretrained CNNs.
\end{itemize}

\subsubsection{Cluster-Aware Grouping for Leakage-Free Splits}
\label{sec:clusters}
To further prevent leakage caused by visually related samples, images were clustered on their pHash values using a Hamming-distance threshold of 4, and each cluster was assigned an identifier so that highly similar images shared the same group. These clusters were used as grouping constraints during data splitting. A total of 48,065 clusters were formed, with a mean size of 1.014 images and a maximum size of 17.

All deduplication and split generation was performed \textit{offline}; the cleaned dataset was then consumed by PyTorch training scripts with no augmentation beyond standard normalization. Cross-validation and cross-dataset evaluation followed the protocols described in Sections~\ref{sec:groupkfold} and~\ref{sec:lodo}.

\subsection{Model Architectures}
\begin{figure}[!htbp]
    \centering
    \includegraphics[width=\linewidth]{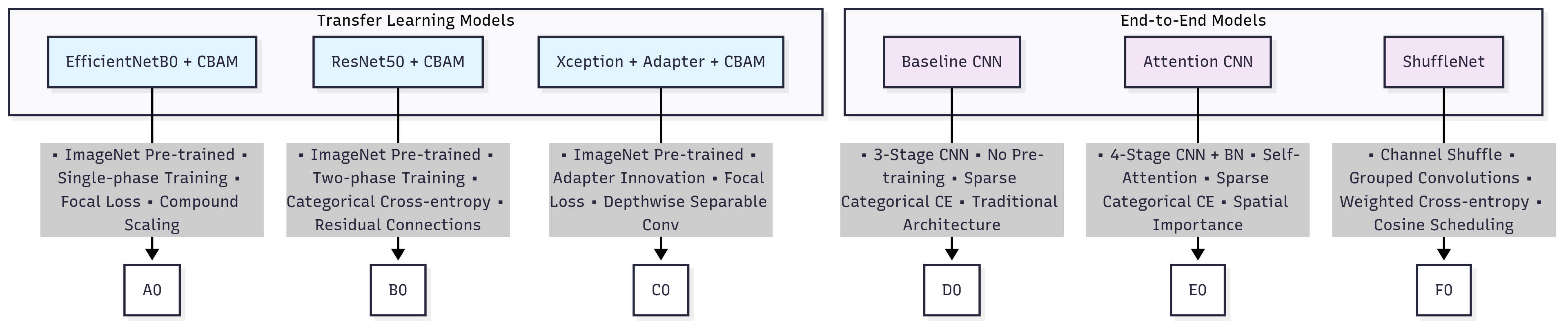}
    \caption{Comparison of the evaluated model architectures.}
    \label{fig:arch-comparison}
\end{figure}

Six deep learning models were evaluated: a baseline CNN, an attention-based CNN, ShuffleNet, EfficientNet-B0, ResNet-50 and Xception. All models follow a common training protocol with model-specific adaptations, summarized in Tables~\ref{tab:training-config} and~\ref{tab:parameters} and Fig.~\ref{fig:arch-comparison}.

\subsubsection{End-to-End Trained Models}
\begin{figure}[!htbp]
    \centering
    \begin{subfigure}[b]{0.3\linewidth}
        \includegraphics[width=\linewidth]{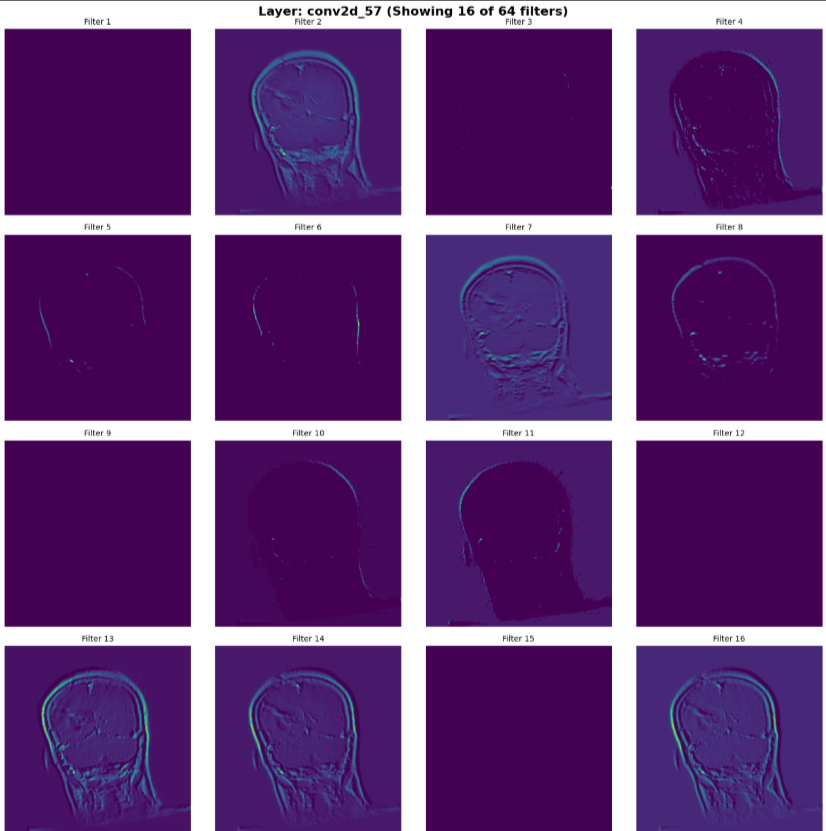}
        \caption{}
    \end{subfigure}\hfill
    \begin{subfigure}[b]{0.3\linewidth}
        \includegraphics[width=\linewidth]{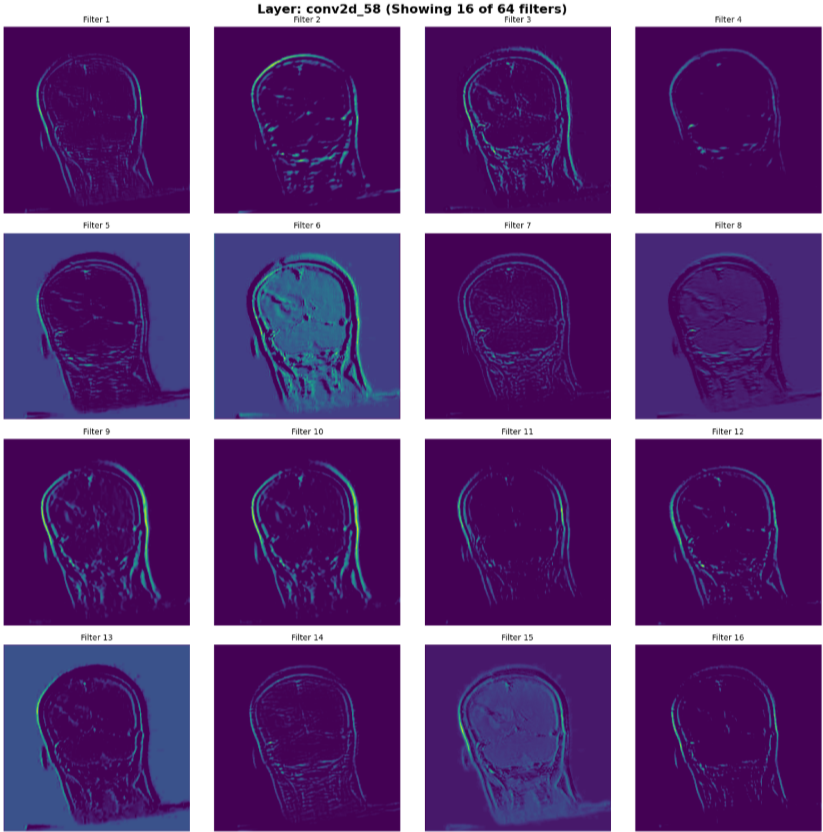}
        \caption{}
    \end{subfigure}\hfill
    \begin{subfigure}[b]{0.3\linewidth}
        \includegraphics[width=\linewidth]{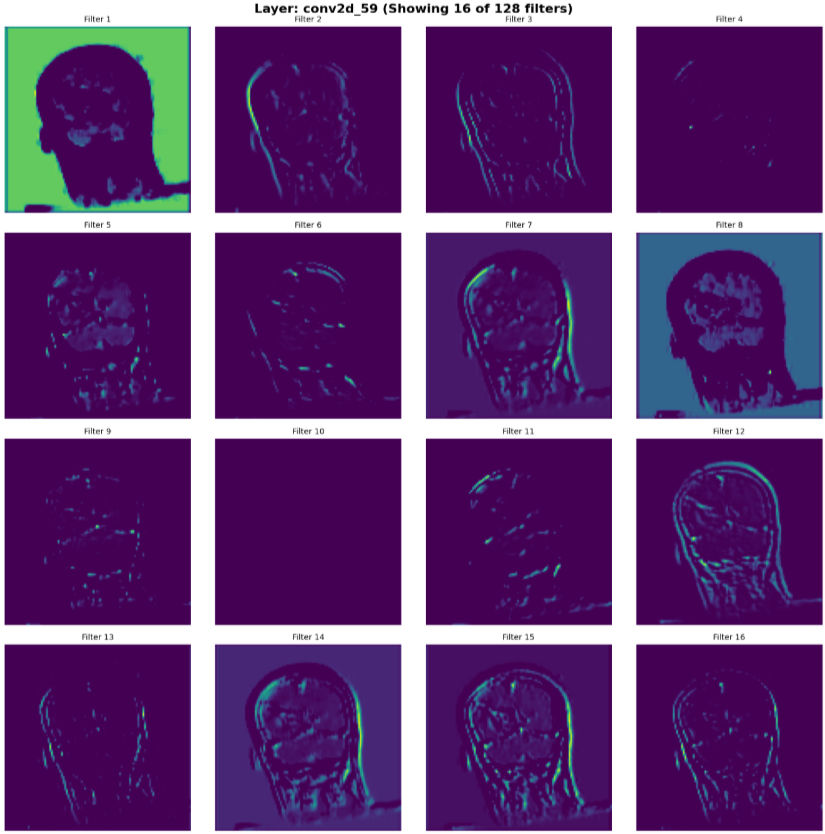}
        \caption{}
    \end{subfigure}

    \medskip
    \begin{subfigure}[b]{0.3\linewidth}
        \includegraphics[width=\linewidth]{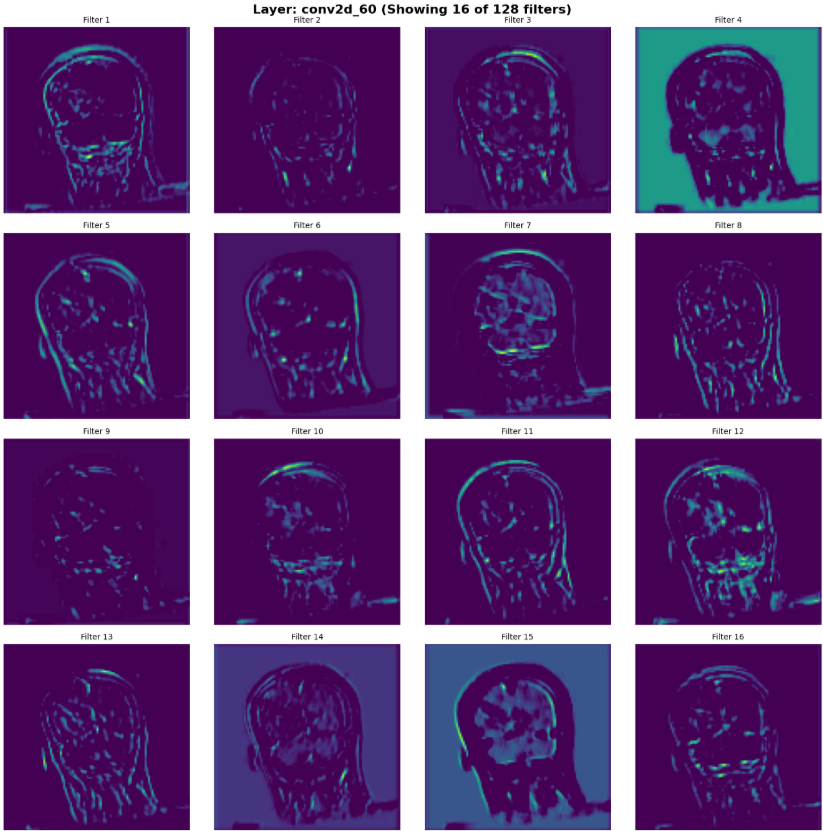}
        \caption{}
    \end{subfigure}\hfill
    \begin{subfigure}[b]{0.3\linewidth}
        \includegraphics[width=\linewidth]{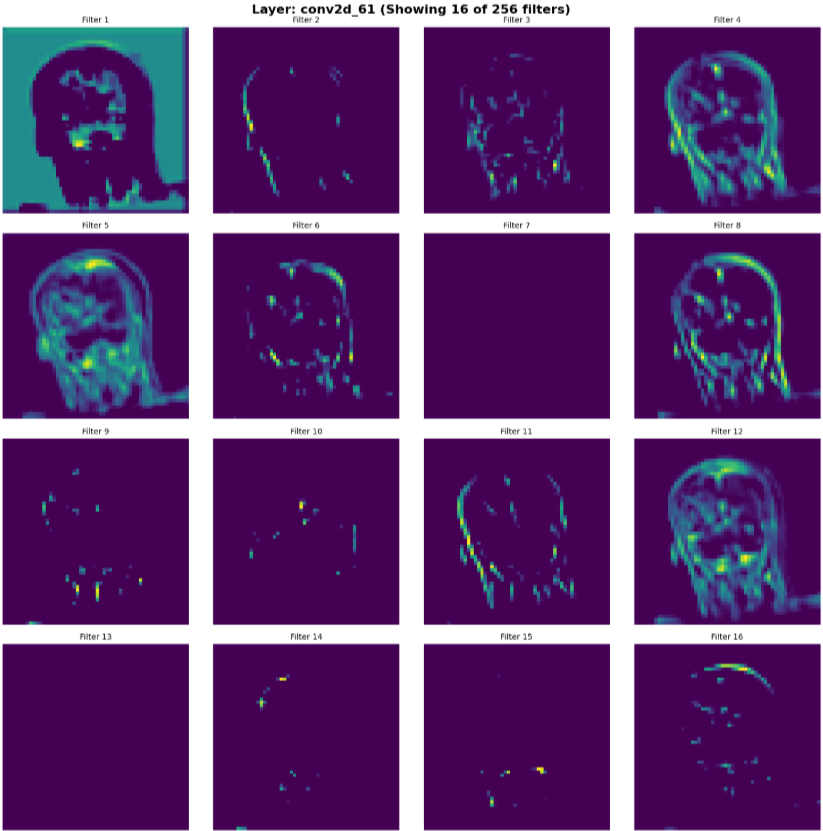}
        \caption{}
    \end{subfigure}\hfill
    \begin{subfigure}[b]{0.3\linewidth}
        \includegraphics[width=\linewidth]{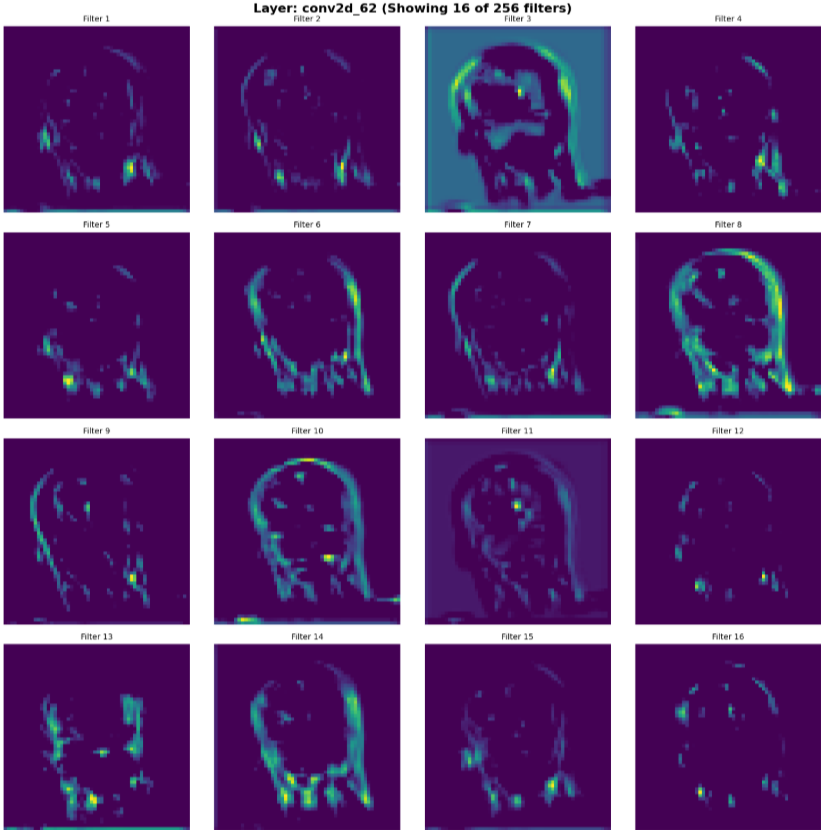}
        \caption{}
    \end{subfigure}
    \caption{Feature maps of an MRI slice at successive convolutional blocks of the CNN: (a), (b) 64-filter block; (c), (d) 128-filter block; (e), (f) 256-filter block.}
    \label{fig:layer-visualization}
\end{figure}

The baseline CNN, the attention CNN and ShuffleNet were initialized from scratch (Xavier/He initialization) and trained directly on the dataset.
\begin{itemize}
    \item \textbf{Baseline CNN:} This model served as the performance benchmark. It used a standard three-stage hierarchical feature extractor (32--64--128 channels) built from sequential blocks of $3\times3$ convolution, ReLU activation and $2\times2$ max pooling (Table~\ref{tab:model-variations}).
    \item \textbf{Attention CNN:} An extension of the baseline model, it used a deeper four-stage convolutional hierarchy (up to 256 channels) with a spatial self-attention mechanism after the final convolutional layer. Feature maps from successive convolutional blocks are shown in Fig.~\ref{fig:layer-visualization}. Its main components are:
    \begin{itemize}
        \item \textbf{Hierarchical feature extraction:} a four-stage CNN produces progressively deeper representations of the input;
        \item \textbf{Spatial attention:} the network learns to weight the importance of different image regions when forming the final feature representation;
        \item \textbf{Context aggregation:} the attention weights produce a context vector that emphasizes discriminative features; and
        \item \textbf{Classification:} a two-layer dense network with dropout produces the final predictions.
    \end{itemize}
    \item \textbf{ShuffleNet:} ShuffleNet uses grouped convolutions and a channel shuffle operation and is optimized for computational efficiency. It reduces the number of floating-point operations (FLOPs) required for inference, making it well suited to resource-constrained environments while maintaining high performance. Its key optimizations are:
    \begin{itemize}
        \item \textbf{Grouped convolutions:} reduce FLOPs by a factor of $g$ (typically $g=3$);
        \item \textbf{Channel shuffling:} enables information exchange between groups;
        \item \textbf{Cosine scheduling:} learning-rate annealing with warm restarts; and
        \item \textbf{Weighted loss:} addresses class imbalance through inverse-frequency weighting.
    \end{itemize}
    This design achieves an approximately eightfold reduction in computation compared with standard convolutions.
\end{itemize}

\begin{table}[!htbp]
\centering
\caption{Architectural differences between the two custom CNNs.}
\label{tab:model-variations}
\resizebox{0.9\columnwidth}{!}{%
\begin{tabular}{|l|l|l|l|l|}
\hline
\multicolumn{1}{|c|}{\textbf{Model}} &
  \multicolumn{1}{c|}{\textbf{Stages}} &
  \multicolumn{1}{c|}{\textbf{Channel progression}} &
  \multicolumn{1}{c|}{\textbf{Spatial reduction}} &
  \multicolumn{1}{c|}{\textbf{Enhancement}} \\ \hline
\textbf{Baseline CNN} &
  3 &
  32$\rightarrow$64$\rightarrow$128 &
  224$\rightarrow$112$\rightarrow$56$\rightarrow$28 &
  None \\ \hline
\textbf{Attention CNN} &
  4 &
  32$\rightarrow$64$\rightarrow$128$\rightarrow$256 &
  224$\rightarrow$112$\rightarrow$56$\rightarrow$28$\rightarrow$14 &
  Batch normalization, self-attention \\ \hline
\end{tabular}%
}
\end{table}

\subsubsection{Transfer Learning and Attention-Enhanced Models}
\begin{figure}[!htbp]
    \centering
    \begin{subfigure}[b]{0.52\linewidth}
        \centering
        \includegraphics[width=\linewidth]{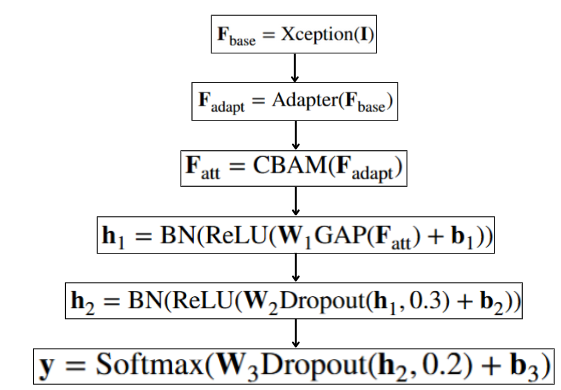}
        \caption{}
        \label{fig:xception-arch}
    \end{subfigure}\hfill
    \begin{subfigure}[b]{0.36\linewidth}
        \centering
        \includegraphics[width=\linewidth]{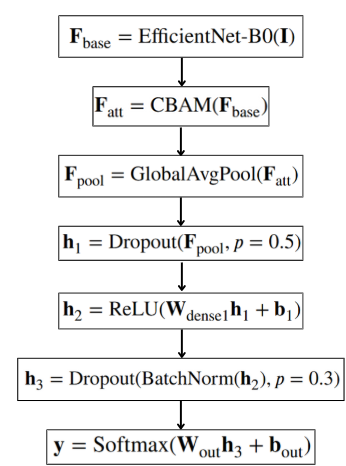}
        \caption{}
        \label{fig:effnet-arch}
    \end{subfigure}
    \caption{Transfer learning architectures: (a) triple-enhanced Xception with adapter and CBAM; (b) EfficientNet-B0 with CBAM integration.}
    \label{fig:transfer-arch}
\end{figure}

For EfficientNet-B0, ResNet-50 and Xception, weights pre-trained on ImageNet were used to accelerate convergence and reuse features learned from a large corpus.
\begin{itemize}
    \item \textbf{CBAM integration:} All three transfer learning models integrate the Convolutional Block Attention Module (CBAM), which refines the feature maps sequentially, applying attention first across channels (\emph{what} is important) and then across spatial dimensions (\emph{where} it is important) before classification.
    \item \textbf{Xception adapter:} The Xception model additionally includes an adapter block, a lightweight bottleneck residual module that refines the pre-trained features and supports efficient domain adaptation with minimal parameter overhead.
    \item \textbf{Triple-enhanced Xception:} The backbone produces features $\mathbf{F}_{\text{base}}$, which the adapter refines into $\mathbf{F}_{\text{adapt}}$ and CBAM enhances into $\mathbf{F}_{\text{att}}$. The classification head then applies batch normalization (BN), ReLU activation, dropout (rates 0.3 and 0.2) and dense layers (Fig.~\ref{fig:xception-arch}).
    \item \textbf{EfficientNet backbone:} EfficientNet-B0 uses compound scaling (depth $\alpha^\phi$, width $\beta^\phi$, resolution $\gamma^\phi$) with mobile inverted bottleneck convolution (MBConv) blocks (Fig.~\ref{fig:effnet-arch}).
    \item \textbf{Training protocols:} Two training protocols were used to adapt the models to deep features. Single-phase training was used for EfficientNet-B0 and Xception, and two-phase training for ResNet-50 only.
    \begin{itemize}
        \item \textbf{Single-phase:} the backbone layers were frozen, and only the CBAM block, the adapter block (Xception only) and the classification head were trained.
        \item \textbf{Two-phase:} the first phase trained the new head and CBAM; in the second phase, the final convolutional block (conv5) was unfrozen and fine-tuned with a learning rate reduced by a factor of 10.
    \end{itemize}
\end{itemize}

\begin{table}[!htbp]
\centering
\caption{Common training configuration.}
\label{tab:training-config}
\resizebox{0.75\columnwidth}{!}{%
\begin{tabular}{|c|c|c|}
\hline
\textbf{Parameter}      & \textbf{Value} & \textbf{Description}                       \\ \hline
Random seed             & 42             & For reproducibility across all experiments \\ \hline
Early stopping patience & 3--10 epochs   & Model-specific (see Table~\ref{tab:parameters}) \\ \hline
Batch normalization     & Applied        & All models except the baseline CNN         \\ \hline
Gradient clipping       & 0.5            & ShuffleNet only                            \\ \hline
Validation split        & 20\%           & Consistent across all models               \\ \hline
\end{tabular}%
}
\end{table}

\subsection{Mathematical Formulation}
Training relied on two main mathematical components: the attention mechanisms used for feature refinement and the loss functions used for optimization.

\subsubsection{Attention Mechanisms}
The CBAM block refines an input feature map $F$ in two sequential steps, yielding the refined feature map $F''$.
\begin{enumerate}
    \item \textbf{Channel attention ($M_c$).}
    Channel attention determines which feature channels are most relevant by aggregating spatial information with global average pooling (GAP) and global max pooling (GMP), followed by a shared multi-layer perceptron (MLP):
    \[
    M_c = \sigma \big( \mathrm{MLP}(\mathrm{GAP}(F)) + \mathrm{MLP}(\mathrm{GMP}(F)) \big), \qquad
    F' = M_c \otimes F,
    \]
    where $\sigma$ is the sigmoid function and $\otimes$ denotes element-wise multiplication.
    \item \textbf{Spatial attention ($M_s$).}
    Spatial attention identifies informative spatial locations by applying a $7\times7$ convolution to the concatenation of the channel-wise average-pooled and max-pooled versions of $F'$:
    \[
    M_s = \sigma \big( \mathrm{Conv}^{7 \times 7}([\mathrm{AvgPool}_c(F');\, \mathrm{MaxPool}_c(F')]) \big), \qquad
    F'' = M_s \otimes F'.
    \]
\end{enumerate}
The spatial self-attention mechanism of the attention CNN computes attention weights over the sequence of spatial features $X_{\text{seq}}$:
\begin{equation}
\mathrm{Attention}(X_{\text{seq}}) = \mathrm{Softmax}\big(X_{\text{seq}} W_{a}^{T}\big)\, X_{\text{seq}},
\end{equation}
where $W_a$ is a learned projection matrix.

\subsubsection{Loss Functions}
The models use loss functions tailored to their training objectives and to the class distribution (Table~\ref{tab:parameters}); the choice of loss is important for addressing class imbalance.
\begin{enumerate}
    \item \textbf{Focal loss.}
    Focal loss down-weights well-classified examples and focuses learning on hard examples, which helps with the class imbalance common in medical imaging datasets:
    \begin{equation}
    \mathrm{FL}(p_t) = -\alpha_t(1-p_t)^{\gamma}\log(p_t),
    \end{equation}
    where
    \begin{equation}
    p_t = \begin{cases}
    p & \text{if } y = 1, \\
    1-p & \text{otherwise},
    \end{cases}
    \end{equation}
    $\alpha_t$ is a weighting factor, $\gamma$ is the focusing parameter and $p \in [0,1]$ is the model's estimated probability for the class with label $y = 1$.
    \textit{Used by:} EfficientNet-B0 and Xception.

    \item \textbf{Categorical cross-entropy.}
    The standard loss for multi-class classification with one-hot encoded labels:
    \begin{equation}
    \mathcal{L}_{\text{CE}} = -\frac{1}{N}\sum_{i=1}^{N}\sum_{c=1}^{C}y_{i,c}\log(\hat{y}_{i,c}),
    \end{equation}
    where $N$ is the number of samples, $C$ the number of classes, $y_{i,c}$ the one-hot true label and $\hat{y}_{i,c}$ the predicted probability.
    \textit{Used by:} ResNet-50.

    \item \textbf{Sparse categorical cross-entropy.}
    An equivalent formulation for integer-encoded labels that avoids one-hot conversion:
    \begin{equation}
    \mathcal{L}_{\text{SCE}} = -\frac{1}{N}\sum_{i=1}^{N}\log(\hat{y}_{i,y_i}),
    \end{equation}
    where $y_i \in \{0,1,2,3\}$ is the true class index and $\hat{y}_{i,y_i}$ is the predicted probability of the true class.
    \textit{Used by:} the baseline CNN and the attention CNN.

    \item \textbf{Weighted cross-entropy.}
    Addresses class imbalance by assigning a weight to each class:
    \begin{equation}
    \mathcal{L}_{\text{weighted}} = -\frac{1}{N} \sum_{i=1}^{N} w_{y_i} \log \big( \hat{y}_{i, y_i} \big),
    \end{equation}
    where $w_c = N/(4\,n_c)$ is the weight of class $c$, inversely proportional to its frequency, and $n_c$ is the number of samples in class $c$.
    \textit{Used by:} ShuffleNet.
\end{enumerate}

\subsection{Model Hyperparameters}
All models used the Adam optimizer together with the regularization and scheduling techniques summarized in Table~\ref{tab:parameters}.

\begin{table}[!htbp]
\centering
\caption{Model-specific training parameters.}
\label{tab:parameters}
\resizebox{\linewidth}{!}{%
\begin{tabular}{lcccccc}
\toprule
\textbf{Model} & \textbf{Learning rate} & \textbf{Optimizer} & \textbf{Loss function} & \textbf{Dropout rates} & \textbf{Early stopping} & \textbf{Training phases} \\
\midrule
EfficientNet-B0 & $10^{-4}$ & Adam & Focal loss ($\alpha=0.25, \gamma=2.0$) & 0.5, 0.3 & 3--5 epochs & Single-phase \\
ResNet-50 & $10^{-4} \rightarrow 10^{-5}$ & Adam & Categorical cross-entropy & 0.3, 0.2 & 5 epochs & Two-phase \\
Xception & $10^{-4}$ & Adam & Focal loss ($\alpha=0.25, \gamma=2.0$) & 0.3, 0.2 & 7 epochs & Single-phase \\
Baseline CNN & $10^{-3}$ & Adam & Sparse categorical cross-entropy & 0.5 & 7 epochs & Single-phase \\
Attention CNN & $10^{-3}$ & Adam & Sparse categorical cross-entropy & 0.5, 0.3 & 10 epochs & Single-phase \\
ShuffleNet & $2.5\times 10^{-4}$ & Adam & Weighted cross-entropy & Progressive$^{*}$ & 10 epochs & Cosine scheduling \\
\bottomrule
\end{tabular}%
}

\smallskip
\raggedright\footnotesize $^{*}$Progressive dropout: $0.1 + 0.05 \times$ stage number.
\end{table}

\subsubsection{Optimization and Regularization Strategy}
\begin{itemize}
    \item \textbf{Learning rate:} Transfer learning models started with a conservative learning rate ($\sim10^{-4}$) to preserve the pre-trained weights, whereas models trained from scratch used a higher learning rate ($\sim10^{-3}$). ResNet-50 reduced its learning rate by a factor of 10 during the fine-tuning phase.
    \item \textbf{Adaptive scheduling:} All experiments used the ReduceLROnPlateau scheduler, which monitors the validation loss and lowers the learning rate when the loss stops improving for a predefined number of epochs.
    \item \textbf{Early stopping:} To limit overfitting, training was stopped when the validation loss failed to improve within the specified patience window.
    \item \textbf{Dropout:} Dropout (rates from 0.2 to 0.5) was applied to the dense layers of the classification heads to improve generalization.
    \item \textbf{Model-specific optimizations:} ShuffleNet used cosine learning-rate annealing and gradient clipping to maintain training stability.
\end{itemize}

\subsection{Evaluation Metrics}
Given the high clinical cost of missed tumors, sensitivity (recall) was prioritized over overall accuracy. Standard classification metrics were selected to capture both a model's ability to detect tumors and the reliability of its positive predictions.

\subsubsection{Overall Accuracy}
Accuracy is the ratio of correctly classified samples to the total number of samples:
\begin{equation}
\text{Accuracy} = \frac{TP + TN}{TP + TN + FP + FN}.
\end{equation}
Accuracy is reported for completeness; because of class imbalance across tumor subtypes, it was not used as the primary indicator.

\subsubsection{Precision, Recall and F1-Score}
In multi-class medical classification, precision, recall and F1-score give a more detailed view of performance for each tumor type.
\begin{itemize}
    \item \textbf{Precision} (positive predictive value) is the ratio of correctly predicted positive observations to all predicted positives. It quantifies the reliability of positive predictions, that is, the extent to which predicted tumor cases correspond to true findings, which matters for avoiding unnecessary clinical follow-up and patient anxiety:
    \begin{equation}
    \text{Precision} = \frac{TP}{TP + FP}.
    \end{equation}
    \item \textbf{Recall} (sensitivity) is the ratio of correctly predicted positive observations to all observations in the actual class. It is the most critical metric here, because false negatives can lead to missed or delayed diagnosis of brain tumors, which can prove fatal:
    \begin{equation}
    \text{Recall} = \frac{TP}{TP + FN}.
    \end{equation}
    \item \textbf{F1-score} is the harmonic mean of precision and recall and is particularly useful for comparing models under class imbalance:
    \begin{equation}
    \text{F1} = 2 \times \frac{\text{Precision} \times \text{Recall}}{\text{Precision} + \text{Recall}}.
    \end{equation}
\end{itemize}

\subsubsection{Area Under the ROC Curve (AUC)}
The area under the receiver operating characteristic (ROC) curve was used to evaluate each model's ability to discriminate between classes across all classification thresholds.

\subsubsection{Precision--Recall Curve}
In addition to ROC-AUC, precision--recall (PR) curves were analyzed for each tumor class. The area under the PR curve gives a more informative assessment on imbalanced medical datasets because it reflects the trade-off between detection sensitivity and false-positive burden.

\section{Results and Discussion}
\subsection{Performance under the Standard Split}
\label{sec:standard-results}
Table~\ref{tab:unified-performance} presents the performance of all six models under the standard train--validation--test split (Section~\ref{sec:standard-pipeline}). The results show substantial differences in classification accuracy, computational efficiency and model complexity, with ShuffleNet achieving the highest overall performance. Under this split, the architectures trained from scratch outperformed the transfer learning models. Note that the test-set sizes differ between models (Table~\ref{tab:classification-report}), so these results are not strictly paired comparisons.

\begin{table}[!htbp]
\centering
\caption{Performance comparison of all models under the standard split.}
\label{tab:unified-performance}
\resizebox{\columnwidth}{!}{%
\begin{tabular}{|l|c|c|c|c|c|c|c|}
\hline
\textbf{Model} & \textbf{Overall accuracy (\%)} & \textbf{Precision} & \textbf{Recall} & \textbf{F1-score} & \textbf{AUC} & \textbf{Parameters (M)} & \textbf{Training epochs} \\ \hline
\textbf{ShuffleNet}      & 97.1 & 0.971 & 0.971 & 0.971 & 0.999 & 1.2  & 35 \\ \hline
\textbf{Attention CNN}   & 95.1 & 0.953 & 0.951 & 0.951 & 0.995 & 2.8  & 40 \\ \hline
\textbf{Baseline CNN}    & 91.2 & 0.912 & 0.912 & 0.912 & 0.989 & 1.5  & 25 \\ \hline
\textbf{Xception}        & 91.0 & 0.913 & 0.910 & 0.910 & 0.980 & 22.9 & 25 \\ \hline
\textbf{EfficientNet-B0} & 88.5 & 0.885 & 0.885 & 0.875 & 0.950 & 5.3  & 20 \\ \hline
\textbf{ResNet-50}       & 84.2 & 0.810 & 0.823 & 0.815 & 0.950 & 25.6 & 30 \\ \hline
\end{tabular}%
}
\end{table}

\subsubsection{Model Training and Convergence}
\begin{figure}[!htbp]
    \centering
    \begin{subfigure}[b]{0.48\linewidth}
        \includegraphics[width=\linewidth]{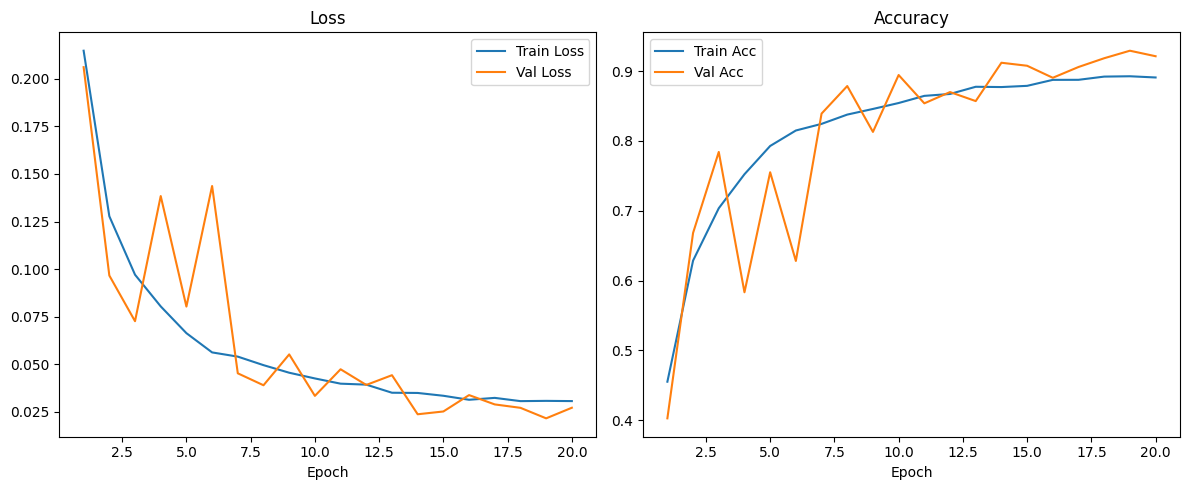}
        \caption{}
    \end{subfigure}\hfill
    \begin{subfigure}[b]{0.48\linewidth}
        \includegraphics[width=\linewidth]{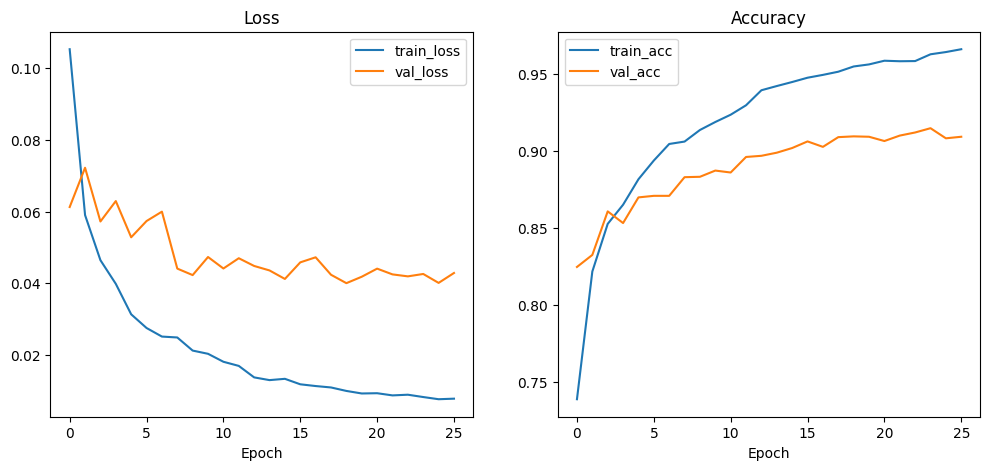}
        \caption{}
    \end{subfigure}

    \medskip
    \begin{subfigure}[b]{0.48\linewidth}
        \includegraphics[width=\linewidth]{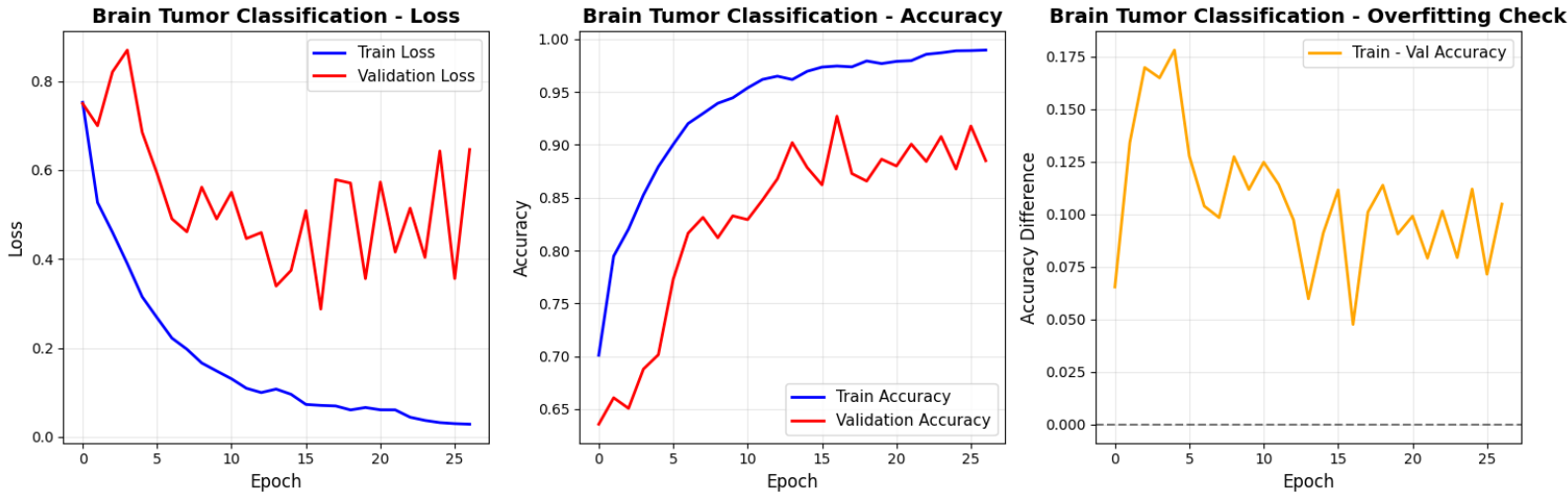}
        \caption{}
    \end{subfigure}\hfill
    \begin{subfigure}[b]{0.48\linewidth}
        \includegraphics[width=\linewidth]{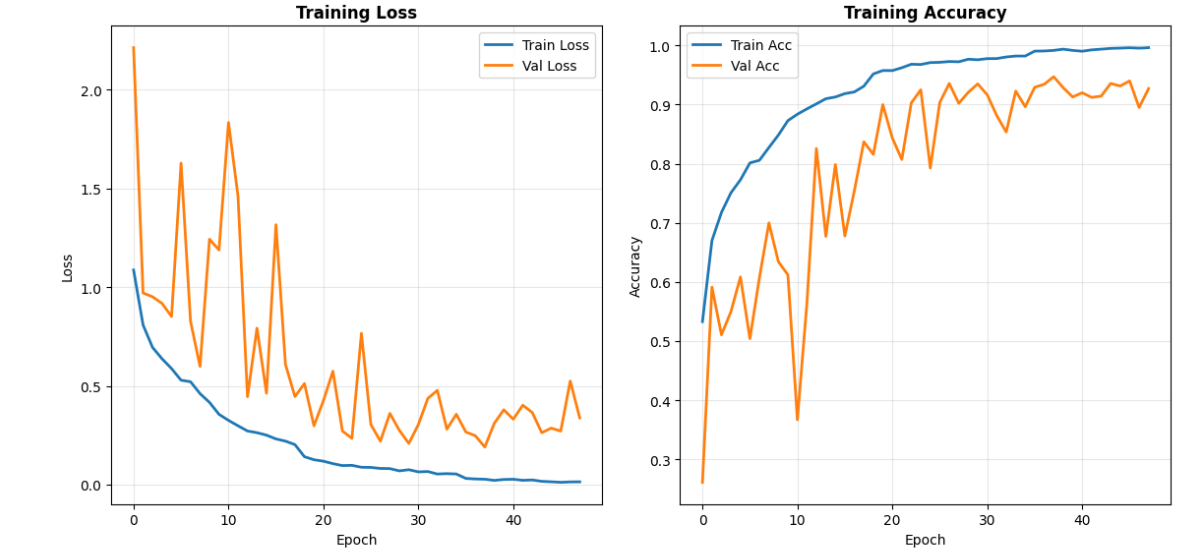}
        \caption{}
    \end{subfigure}

    \medskip
    \begin{subfigure}[b]{0.6\linewidth}
        \includegraphics[width=\linewidth]{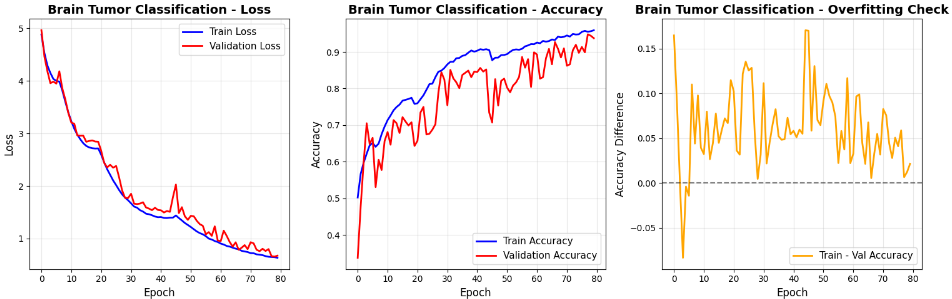}
        \caption{}
    \end{subfigure}
    \caption{Training and validation loss and accuracy curves: (a) EfficientNet-B0, (b) Xception, (c) baseline CNN, (d) attention CNN, (e) ShuffleNet.}
    \label{fig:loss-accuracy}
\end{figure}

Training behavior varied substantially across architectures (Fig.~\ref{fig:loss-accuracy}).
\begin{itemize}
    \item \textbf{Models trained from scratch:} ShuffleNet showed the most stable convergence (Fig.~\ref{fig:loss-accuracy}e), with loss decreasing smoothly from 5.0 to below 0.5 over 35 epochs and training and validation curves closely aligned. It reached approximately 97\% accuracy and a recall of 95.1\% with minimal overfitting. The attention CNN showed improved stability and achieved an overall test accuracy of 95.1\%, with validation-loss fluctuations diminishing after epoch 20 (Fig.~\ref{fig:loss-accuracy}d). The baseline CNN underperformed the other two, achieving an accuracy of 91.2\% (Fig.~\ref{fig:loss-accuracy}c).
    \item \textbf{Transfer learning models:} EfficientNet-B0 (Fig.~\ref{fig:loss-accuracy}a) converged smoothly with minimal overfitting; its training loss decreased from approximately 0.200 to about 0.025 over 20 epochs without significant oscillation. ResNet-50 showed more volatile training and required careful learning-rate scheduling across its two phases. Xception (Fig.~\ref{fig:loss-accuracy}b) converged smoothly and consistently over 25 epochs: training accuracy rose steadily from 74\% to 96\%, while validation accuracy reached approximately 91\% with little fluctuation.
\end{itemize}

\subsubsection{Computational Efficiency versus Performance}
The relationship between model complexity and performance has direct implications for clinical deployment. ShuffleNet achieved the best balance, reaching 97.1\% accuracy with only 1.2M parameters, whereas ResNet-50 required 25.6M parameters to reach 84.2\%. This 21-fold difference in parameter count shows that, under this split, an efficient architecture outperformed a much larger one.

The grouped convolutions and channel shuffling in ShuffleNet reduce FLOPs by a factor of three. This efficiency makes deployment feasible in resource-constrained clinical environments, including point-of-care devices and telemedicine applications, where high-performance computing infrastructure may be unavailable.

\subsubsection{Class-Specific Performance}
\label{sec:class-specific}
Table~\ref{tab:classification-report} presents per-class performance, which shows consistent patterns across all models. Healthy scans (ShuffleNet recall: 99.4\%) and pituitary tumors (ShuffleNet recall: 98.5\%) were the easiest classes for every model, owing to their distinct anatomical and imaging characteristics.

\begin{figure}[!htbp]
    \centering
    \begin{subfigure}[b]{0.3\linewidth}
        \includegraphics[width=\linewidth]{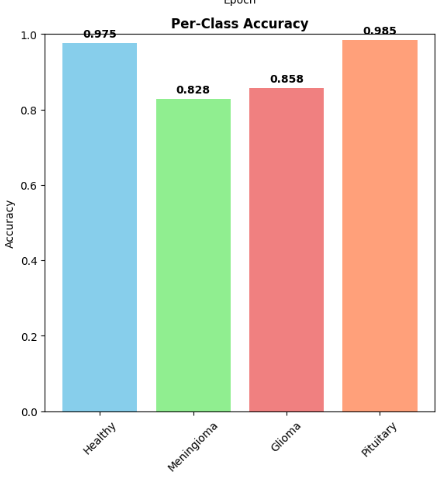}
        \caption{}
    \end{subfigure}\hfill
    \begin{subfigure}[b]{0.3\linewidth}
        \includegraphics[width=\linewidth]{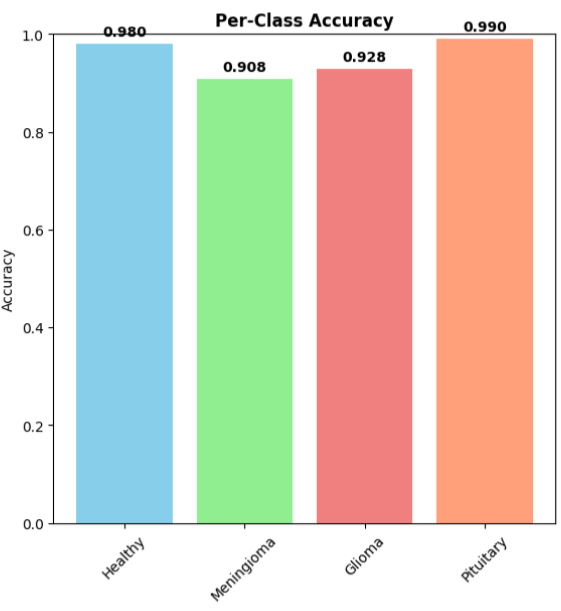}
        \caption{}
    \end{subfigure}\hfill
    \begin{subfigure}[b]{0.3\linewidth}
        \includegraphics[width=\linewidth]{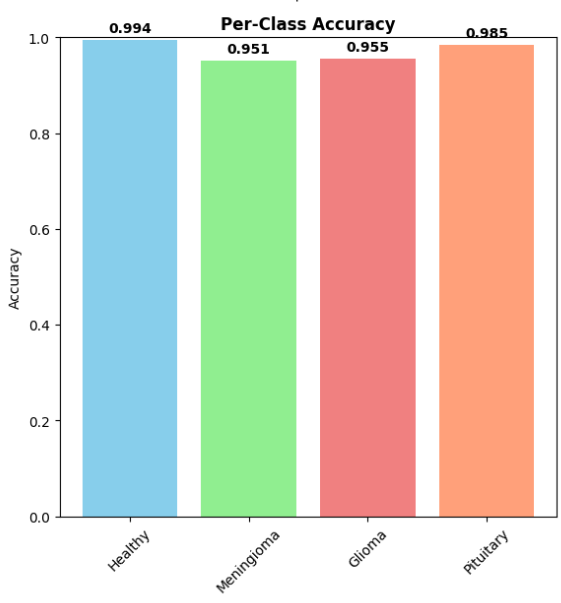}
        \caption{}
    \end{subfigure}
    \caption{Class-wise accuracy: (a) baseline CNN, (b) attention CNN, (c) ShuffleNet.}
    \label{fig:perclass}
\end{figure}

In contrast, glioma and meningioma were the most challenging classes, with the greatest confusion occurring between the two. This is likely due to overlapping features, similar contrast-enhancement patterns and shared locations in the cerebral hemispheres.
The 4--7\% improvement in glioma--meningioma discrimination provided by the attention mechanism suggests that spatial attention captures subtle morphological differences that the plain CNN misses. Class-wise accuracies for the custom models are shown in Fig.~\ref{fig:perclass}.

\begin{table}[!htbp]
\centering
\caption{Per-class classification report under the standard split.}
\label{tab:classification-report}
\resizebox{0.85\columnwidth}{!}{%
\begin{tabular}{|l|l|c|c|c|c|}
\hline
\textbf{Model} & \textbf{Tumor type} & \textbf{Precision} & \textbf{Recall} & \textbf{F1-score} & \textbf{Support} \\ \hline
\multirow{4}{*}{\textbf{EfficientNet-B0}} & Meningioma & 0.80  & 0.83 & 0.81  & 930  \\ \cline{2-6}
                                        & Glioma     & 0.94  & 0.80 & 0.87  & 1208 \\ \cline{2-6}
                                        & Pituitary  & 0.88  & 0.96 & 0.92  & 992  \\ \cline{2-6}
                                        & Healthy    & 0.92  & 0.97 & 0.92  & 992  \\ \hline
\multirow{4}{*}{\textbf{ResNet-50}}     & Meningioma & 0.72  & 0.64 & 0.68  & 930  \\ \cline{2-6}
                                        & Glioma     & 0.81  & 0.78 & 0.79  & 1208 \\ \cline{2-6}
                                        & Pituitary  & 0.88  & 0.90 & 0.89  & 992  \\ \cline{2-6}
                                        & Healthy    & 0.83  & 0.97 & 0.90  & 831  \\ \hline
\multirow{4}{*}{\textbf{Xception}}      & Meningioma & 0.82  & 0.89 & 0.85  & 930  \\ \cline{2-6}
                                        & Glioma     & 0.93  & 0.84 & 0.88  & 1208 \\ \cline{2-6}
                                        & Pituitary  & 0.93  & 0.96 & 0.95  & 992  \\ \cline{2-6}
                                        & Healthy    & 0.97  & 0.99 & 0.98  & 831  \\ \hline
\multirow{4}{*}{\textbf{Baseline CNN}}  & Meningioma & 0.866 & 0.967 & 0.960 & 1000 \\ \cline{2-6}
                                        & Glioma     & 0.918 & 0.883 & 0.900 & 1000 \\ \cline{2-6}
                                        & Pituitary  & 0.910 & 0.961 & 0.935 & 1000 \\ \cline{2-6}
                                        & Healthy    & 0.954 & 0.967 & 0.960 & 1000 \\ \hline
\multirow{4}{*}{\textbf{Attention CNN}} & Meningioma & 0.950 & 0.908 & 0.928 & 1000 \\ \cline{2-6}
                                        & Glioma     & 0.977 & 0.928 & 0.952 & 1000 \\ \cline{2-6}
                                        & Pituitary  & 0.905 & 0.990 & 0.946 & 1000 \\ \cline{2-6}
                                        & Healthy    & 0.980 & 0.980 & 0.980 & 1000 \\ \hline
\multirow{4}{*}{\textbf{ShuffleNet}}    & Meningioma & 0.966 & 0.951 & 0.958 & 800  \\ \cline{2-6}
                                        & Glioma     & 0.961 & 0.955 & 0.958 & 800  \\ \cline{2-6}
                                        & Pituitary  & 0.969 & 0.985 & 0.977 & 800  \\ \cline{2-6}
                                        & Healthy    & 0.989 & 0.994 & 0.991 & 800  \\ \hline
\end{tabular}%
}
\end{table}

\subsubsection{Inter-Tumor Discrimination}
\begin{figure}[!htbp]
    \centering
    \begin{subfigure}[b]{0.31\linewidth}
        \includegraphics[width=\linewidth]{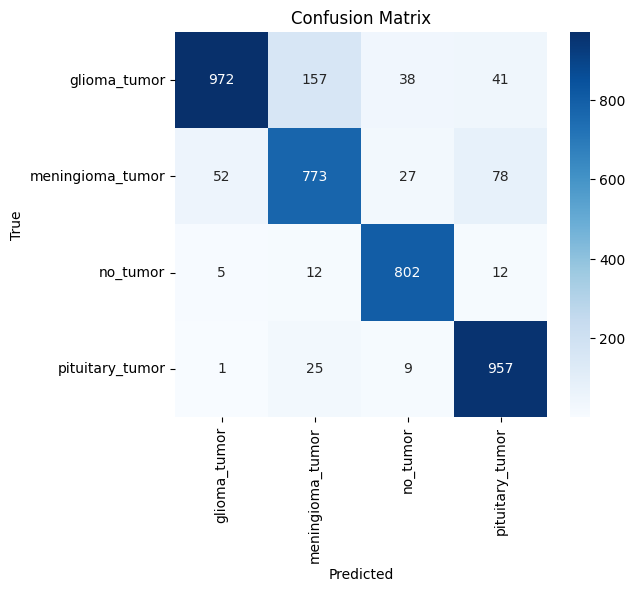}
        \caption{}
    \end{subfigure}\hfill
    \begin{subfigure}[b]{0.31\linewidth}
        \includegraphics[width=\linewidth]{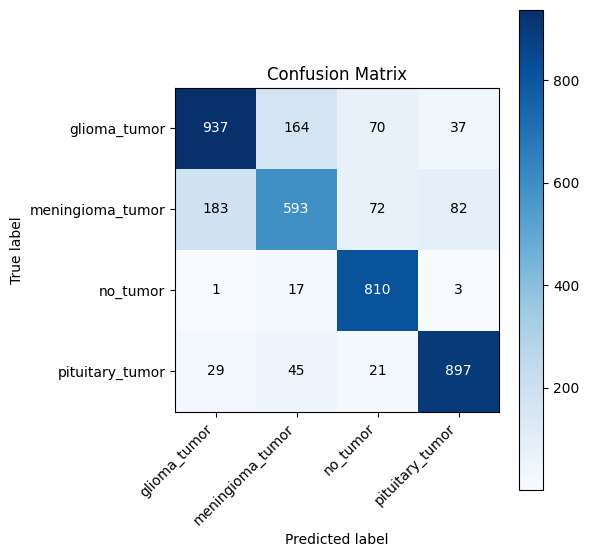}
        \caption{}
    \end{subfigure}\hfill
    \begin{subfigure}[b]{0.31\linewidth}
        \includegraphics[width=\linewidth]{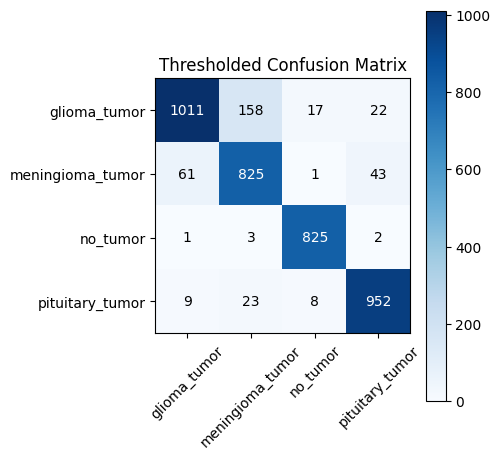}
        \caption{}
    \end{subfigure}

    \medskip
    \begin{subfigure}[b]{0.31\linewidth}
        \includegraphics[width=\linewidth]{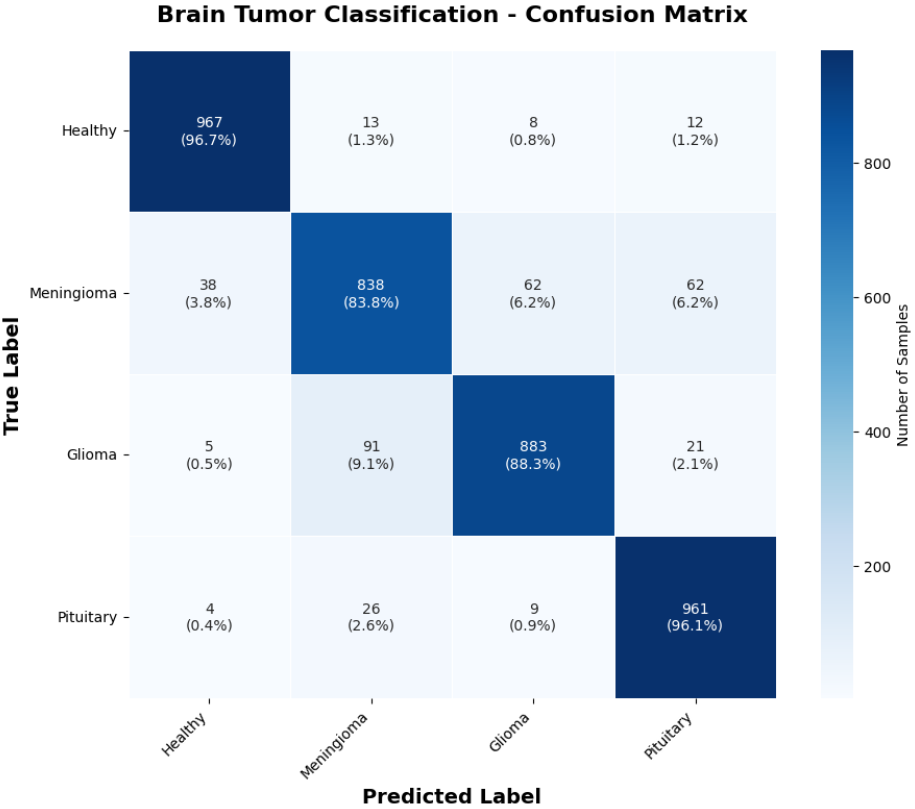}
        \caption{}
    \end{subfigure}\hfill
    \begin{subfigure}[b]{0.31\linewidth}
        \includegraphics[width=\linewidth]{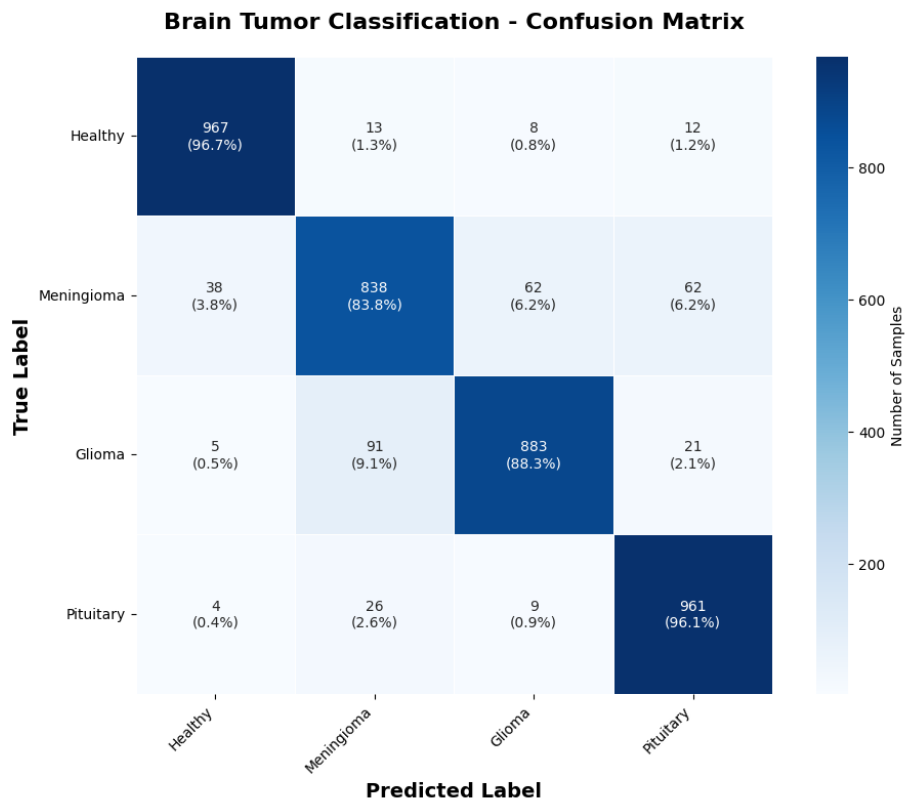}
        \caption{}
    \end{subfigure}\hfill
    \begin{subfigure}[b]{0.31\linewidth}
        \includegraphics[width=\linewidth]{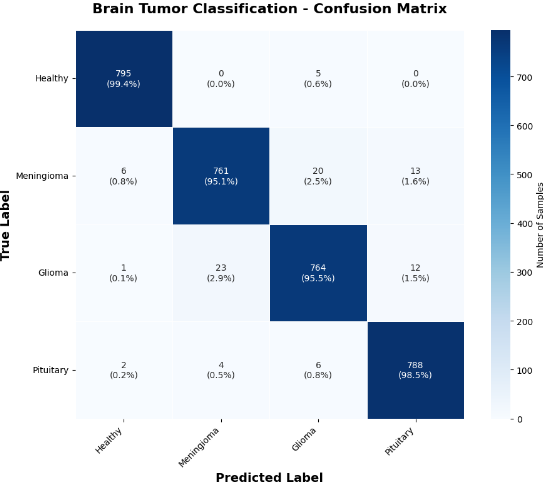}
        \caption{}
    \end{subfigure}
    \caption{Confusion matrices under the standard split: (a) EfficientNet-B0, (b) ResNet-50, (c) Xception, (d) baseline CNN, (e) attention CNN, (f) ShuffleNet.}
    \label{fig:confusion}
\end{figure}

The confusion matrices (Fig.~\ref{fig:confusion}) show that the most frequent errors occurred between the glioma and meningioma classes; the baseline CNN misclassified 91 glioma cases as meningioma, consistent with the analysis in Section~\ref{sec:class-specific}. ShuffleNet discriminated best, correctly classifying 795 healthy cases (99.4\%), 761 meningioma cases (95.1\%), 764 glioma cases (95.5\%) and 789 pituitary cases (98.5\%). Its most notable errors were 20 meningioma cases classified as glioma (2.5\%) and 23 glioma cases classified as meningioma (2.9\%).

The effect of the attention mechanism is evident when comparing the baseline and attention CNNs. Both models performed similarly on healthy tissue and pituitary tumors, but the attention model showed marked improvements in glioma (88.3\%) and meningioma (83.8\%) discrimination.

\subsubsection{ROC and Precision--Recall Analysis}
\begin{figure}[!htbp]
    \centering
    \begin{subfigure}[b]{0.31\linewidth}
        \includegraphics[width=\linewidth]{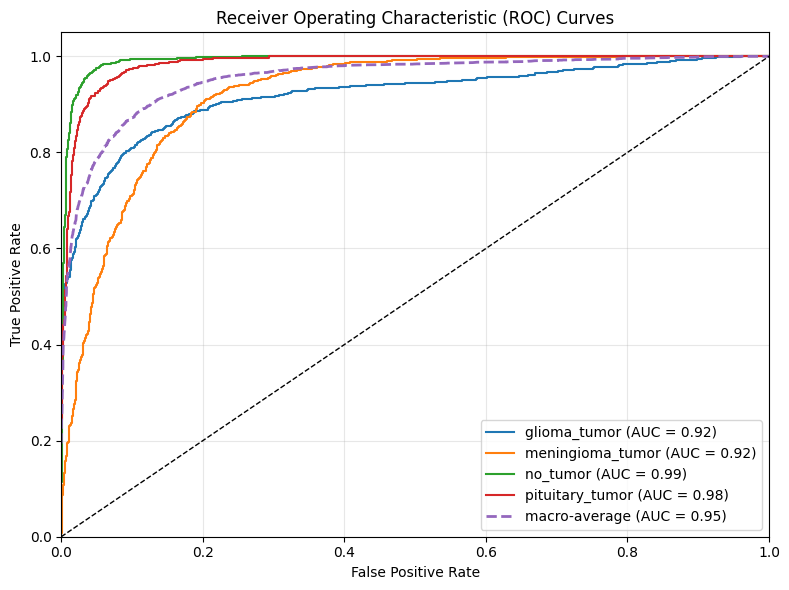}
        \caption{}
    \end{subfigure}\hfill
    \begin{subfigure}[b]{0.31\linewidth}
        \includegraphics[width=\linewidth]{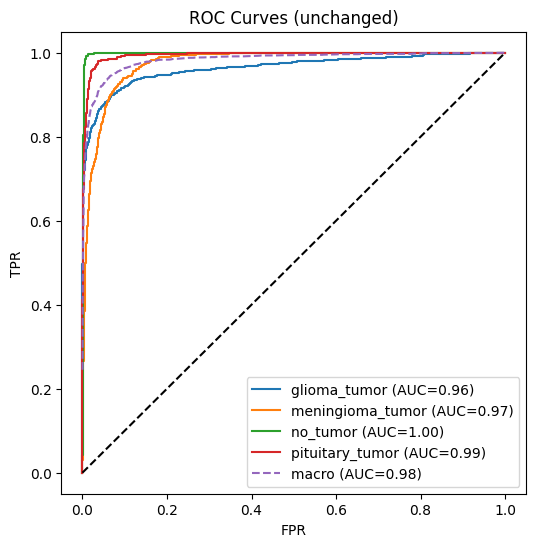}
        \caption{}
    \end{subfigure}\hfill
    \begin{subfigure}[b]{0.31\linewidth}
        \includegraphics[width=\linewidth]{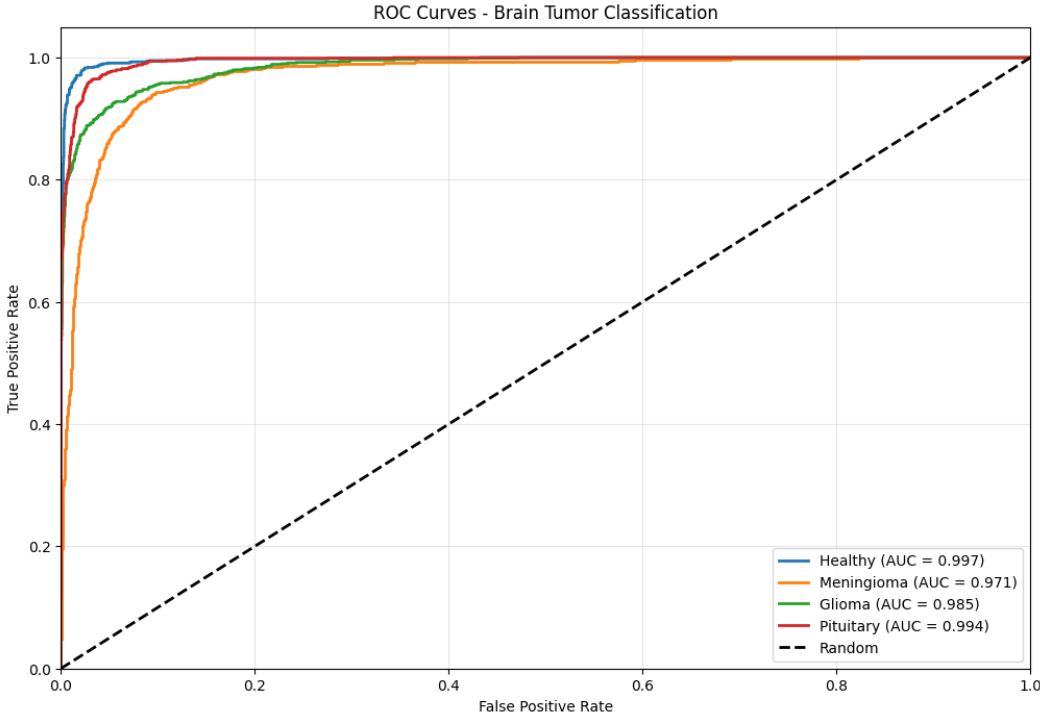}
        \caption{}
    \end{subfigure}

    \medskip
    \begin{subfigure}[b]{0.31\linewidth}
        \includegraphics[width=\linewidth]{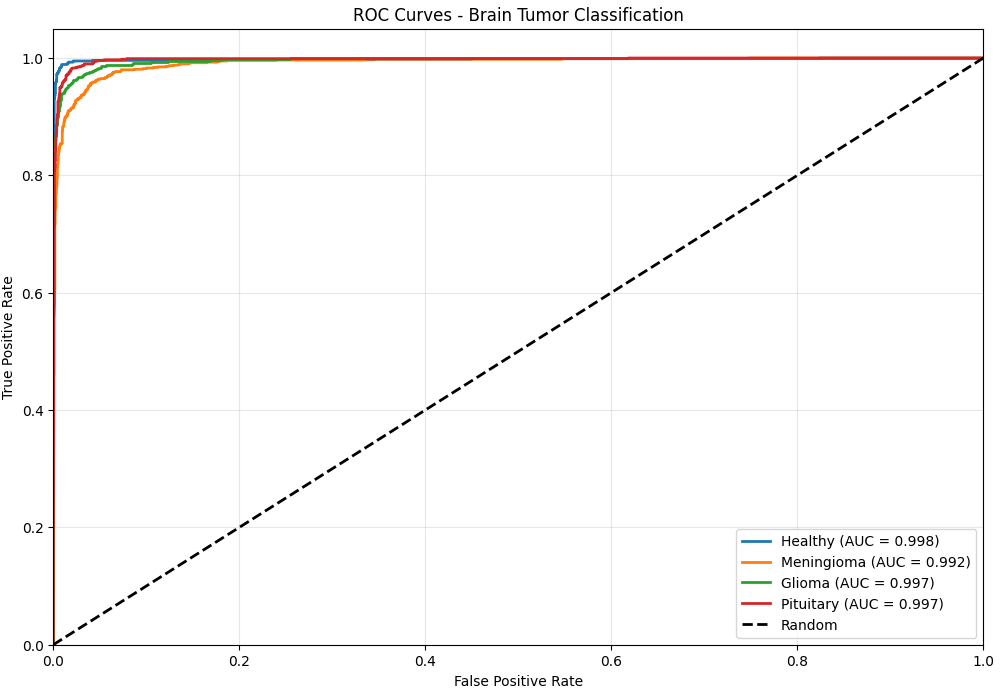}
        \caption{}
    \end{subfigure}\hspace{0.03\linewidth}
    \begin{subfigure}[b]{0.31\linewidth}
        \includegraphics[width=\linewidth]{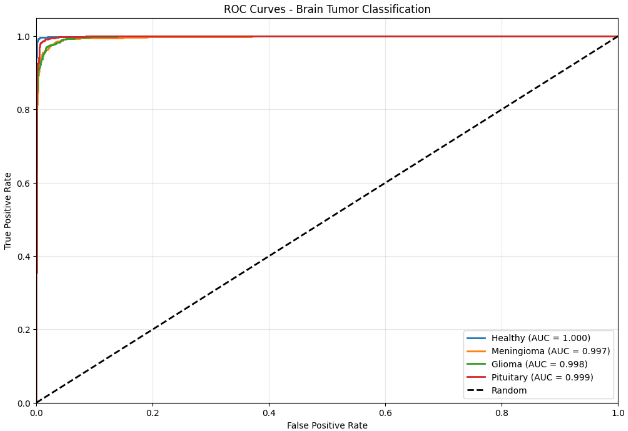}
        \caption{}
    \end{subfigure}
    \caption{ROC curves: (a) ResNet-50, (b) Xception, (c) baseline CNN, (d) attention CNN, (e) ShuffleNet.}
    \label{fig:roc}
\end{figure}

ROC analysis (Fig.~\ref{fig:roc}) showed strong discriminative capability across all models, with AUC values of at least 0.95. ShuffleNet achieved near-perfect discrimination, with per-class AUC values close to 1.0 (healthy: 1.000, meningioma: 0.997, glioma: 0.998, pituitary: 0.999), followed closely by the attention CNN, whose per-class AUC values ranged from 0.992 to 0.998.
AUC values above 0.99 indicate that an operating point exists at which both the false-positive rate and the false-negative rate are low. Low false-positive rates are important for clinical acceptance, and low false-negative rates minimize missed diagnoses, which is critical for patient safety.

\begin{figure}[!htbp]
    \centering
    \begin{subfigure}[b]{0.31\linewidth}
        \includegraphics[width=\linewidth]{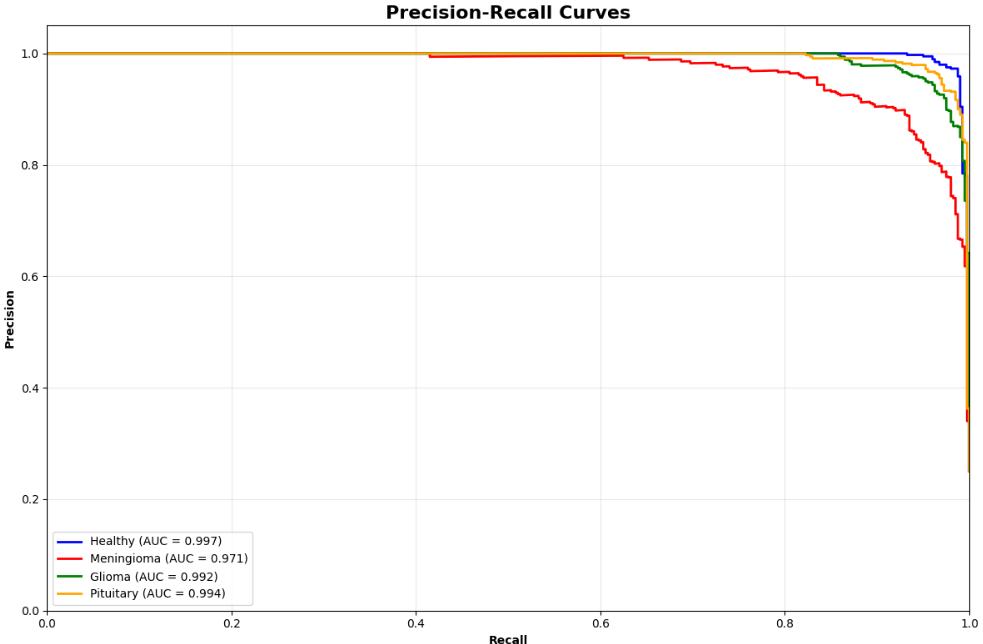}
        \caption{}
    \end{subfigure}\hfill
    \begin{subfigure}[b]{0.31\linewidth}
        \includegraphics[width=\linewidth]{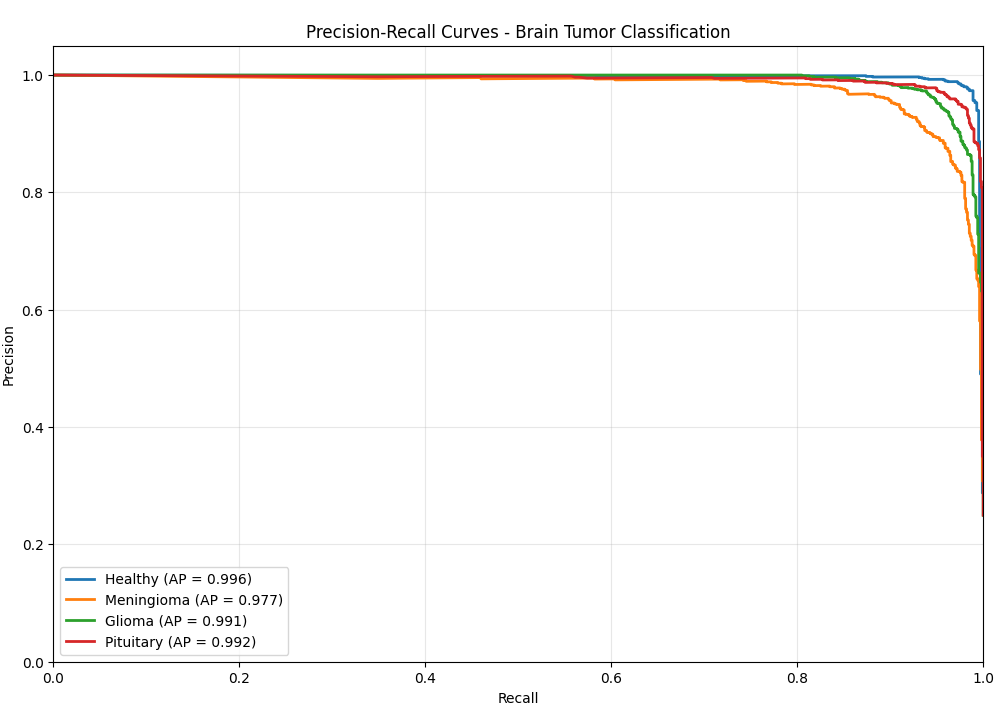}
        \caption{}
    \end{subfigure}\hfill
    \begin{subfigure}[b]{0.31\linewidth}
        \includegraphics[width=\linewidth]{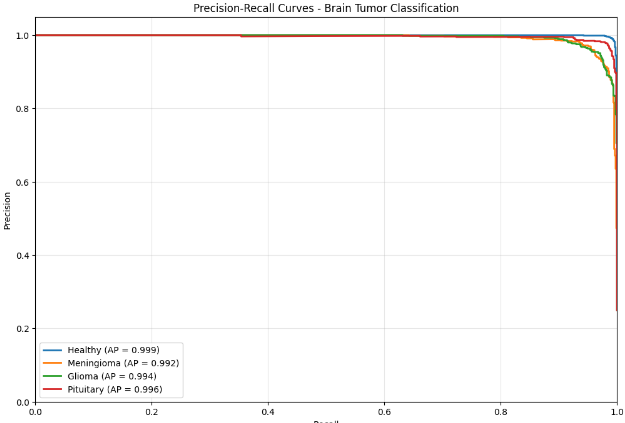}
        \caption{}
    \end{subfigure}
    \caption{Precision--recall curves: (a) baseline CNN, (b) attention CNN, (c) ShuffleNet.}
    \label{fig:pr}
\end{figure}

Precision--recall analysis (Fig.~\ref{fig:pr}) showed that the best-performing models maintained high precision across recall thresholds. ShuffleNet's per-class areas under the PR curve (healthy: 0.999, meningioma: 0.992, glioma: 0.994, pituitary: 0.996) confirm consistent performance across decision thresholds.

\subsubsection{Interpretability of the Attention CNN}
\begin{figure}[!htbp]
    \centering
    \includegraphics[width=0.8\linewidth]{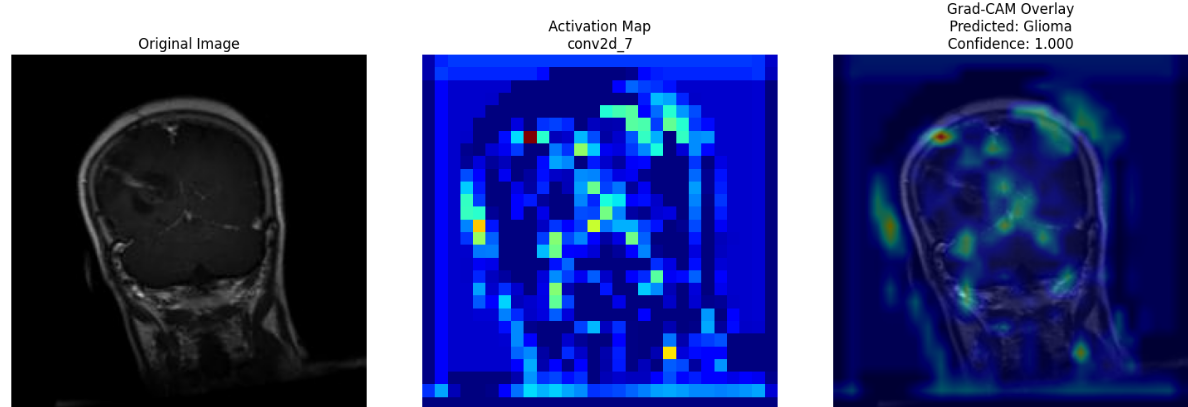}
    \caption{Grad-CAM visualization of the attention CNN.}
    \label{fig:gradcam-attcnn}
\end{figure}

Grad-CAM visualization (Fig.~\ref{fig:gradcam-attcnn}) provides insight into the decision-making of the attention CNN. The activation maps concentrate on tumor-relevant regions and highlight areas of visible pathology; for glioma cases, activation concentrates on irregular tissue boundaries and contrast-enhancing regions.
This indicates that the network learned to focus on relevant regions without explicit localization supervision, which adds transparency to its diagnostic decisions.

\subsubsection{Prediction Confidence}
\begin{figure}[!htbp]
    \centering
    \begin{subfigure}[b]{0.78\linewidth}
        \includegraphics[width=\linewidth]{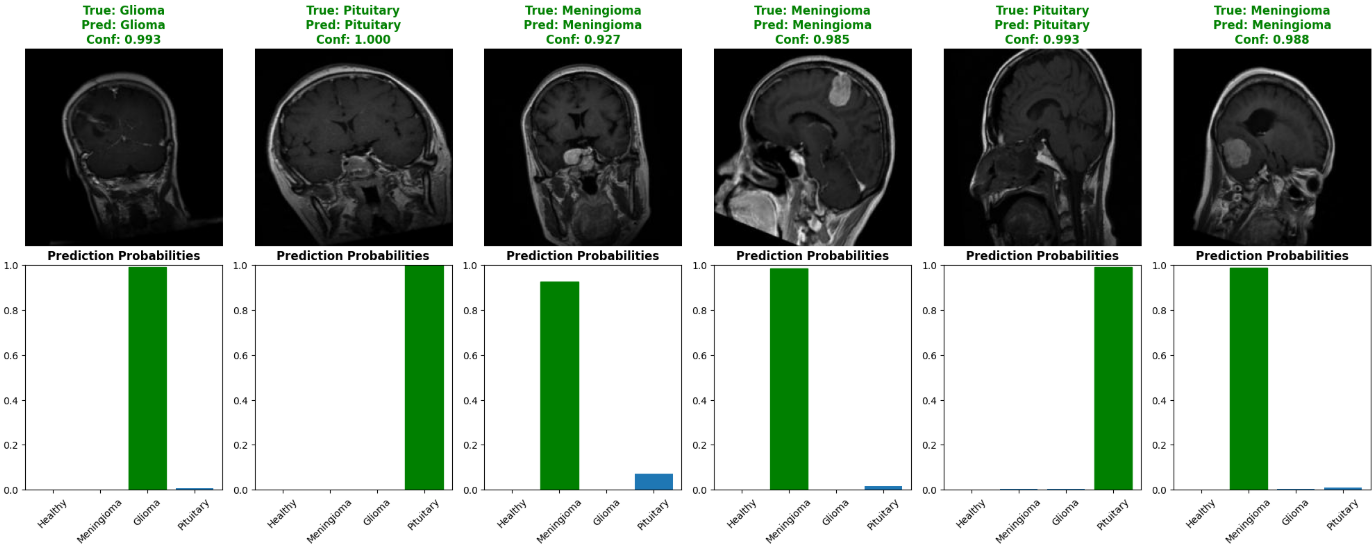}
        \caption{}
    \end{subfigure}

    \begin{subfigure}[b]{0.78\linewidth}
        \includegraphics[width=\linewidth]{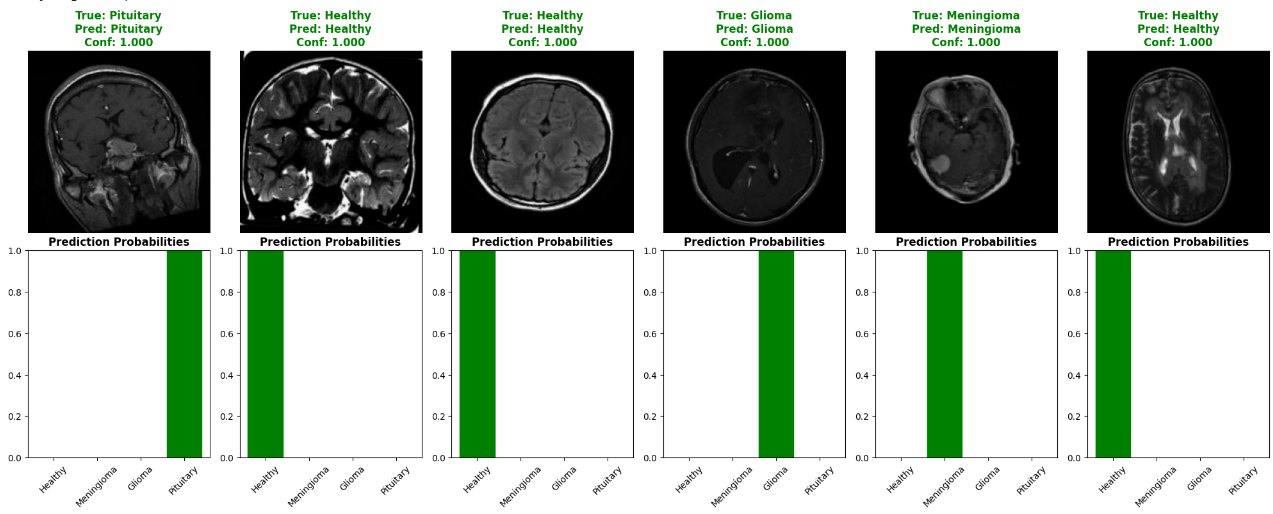}
        \caption{}
    \end{subfigure}

    \begin{subfigure}[b]{0.78\linewidth}
        \includegraphics[width=\linewidth]{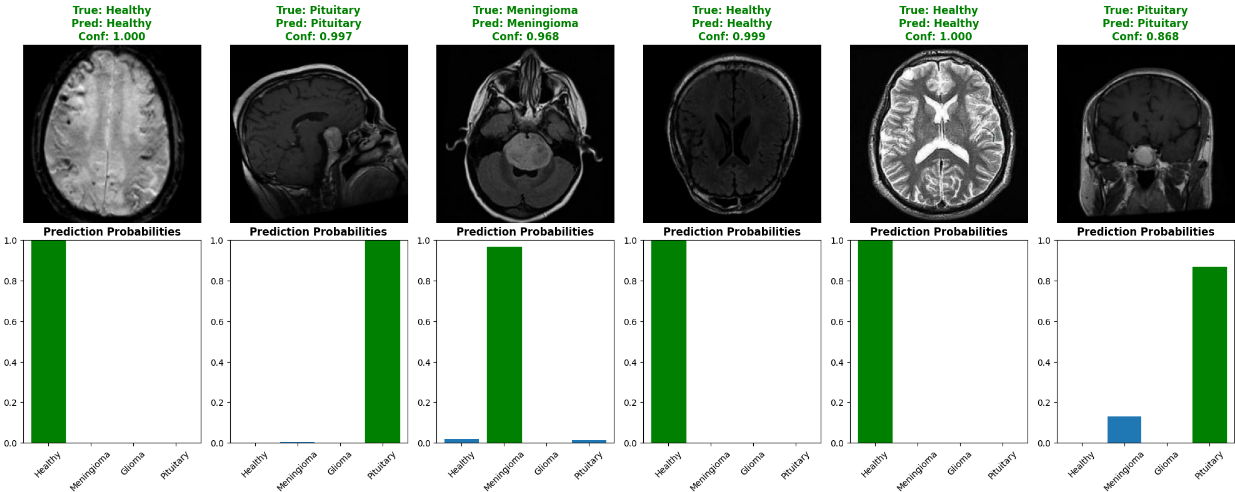}
        \caption{}
    \end{subfigure}
    \caption{Sample predictions with confidence scores: (a) baseline CNN, (b) attention CNN, (c) ShuffleNet.}
    \label{fig:confidence}
\end{figure}

The distributions of predicted class probabilities (Fig.~\ref{fig:confidence}) show that ShuffleNet made most predictions with high confidence (probability above 0.9). Lower confidence scores (0.252--0.292) occurred for challenging cases with ambiguous imaging characteristics.
The attention CNN was more confident than the baseline model: correct classifications received very high confidence scores ($>0.999$), while uncertain cases received lower scores. This better-separated confidence distribution is consistent with the attention CNN's higher accuracy under this split.

\subsubsection{Transfer Learning Performance under the Standard Split}
Under the standard split, the transfer learning models underperformed despite their architectural sophistication. Two factors may explain this:
\begin{itemize}
    \item \textbf{Domain gap:} ImageNet pre-training captures natural-image features (textures, objects, scenes) that may transfer poorly to medical imaging, whereas brain MRI characteristics (tissue contrast, anatomical structures and pathological patterns) call for specialized feature extraction.
    \item \textbf{Task-specific learning:} custom architectures learn medically relevant features from initialization, and attention mechanisms can capture spatial relationships specific to tumor morphology.
\end{itemize}
The poor performance of ResNet-50 (84.2\%) despite its 25.6M parameters suggests that depth alone, without domain-specific design, gives limited benefit under this protocol. As shown in the following sections, however, this ranking does not hold once data leakage is controlled.

\subsection{Performance under Group K-Fold Cross-Validation}
\label{sec:kfold-results}
To evaluate in-distribution performance while strictly avoiding data leakage, all architectures were trained and evaluated with 5-fold group cross-validation on the deduplicated dataset, with pHash clusters kept within folds (Section~\ref{sec:leakage-pipeline}). This ensured that duplicated or visually similar MRI scans were never split between training and validation sets.
Table~\ref{tab:kfold} summarizes the mean, standard deviation and range of accuracy and the mean F1-score across the five folds.

\begin{table}[!htbp]
\centering
\caption{Performance of all models under 5-fold group cross-validation (mean and standard deviation across folds).}
\label{tab:kfold}
\begin{tabular}{lcccccc}
\toprule
\textbf{Model} & \textbf{Mean acc.} & \textbf{Std} & \textbf{Min acc.} & \textbf{Max acc.} & \textbf{Mean F1} & \textbf{Std F1} \\
\midrule
ShuffleNetV2    & 0.9841 & 0.0016 & 0.9818 & 0.9864 & 0.9841 & 0.0016 \\
EfficientNet-B0 & 0.9823 & 0.0021 & 0.9796 & 0.9849 & 0.9824 & 0.0021 \\
Xception        & 0.9741 & 0.0031 & 0.9691 & 0.9784 & 0.9742 & 0.0031 \\
ResNet-50       & 0.9521 & 0.0033 & 0.9469 & 0.9561 & 0.9523 & 0.0033 \\
Baseline CNN    & 0.8950 & 0.0070 & 0.8830 & 0.9036 & 0.8946 & 0.0074 \\
Attention CNN   & 0.8071 & 0.0099 & 0.7943 & 0.8233 & 0.8068 & 0.0102 \\
\bottomrule
\end{tabular}
\end{table}

ShuffleNetV2 achieved the best overall performance, with a mean accuracy of $0.9841 \pm 0.0016$ and an identical mean F1-score of 0.9841, showing both high discriminative power and stability across folds. Its very low standard deviation indicates robustness to data partitioning under leakage-free conditions.

EfficientNet-B0 followed closely with a mean accuracy of $0.9823 \pm 0.0021$, and Xception achieved $0.9741 \pm 0.0031$. Both performed strongly but with slightly higher variance than ShuffleNetV2, suggesting marginally lower stability across folds.

ResNet-50 achieved moderate performance ($0.9521 \pm 0.0033$), while the custom baseline and attention CNNs lagged considerably, with mean accuracies of 0.8950 and 0.8071, respectively. This ranking reverses the finding of Section~\ref{sec:standard-results}, where the transfer learning models appeared to underperform. The two protocols differ in implementation and preprocessing, so the comparison is not exact; however, one plausible explanation is that the standard split, which was not deduplicated, allowed near-identical images to appear in both training and test sets, favoring models that memorize the training distribution.

\begin{figure}[!htbp]
    \centering
    \includegraphics[width=0.8\linewidth]{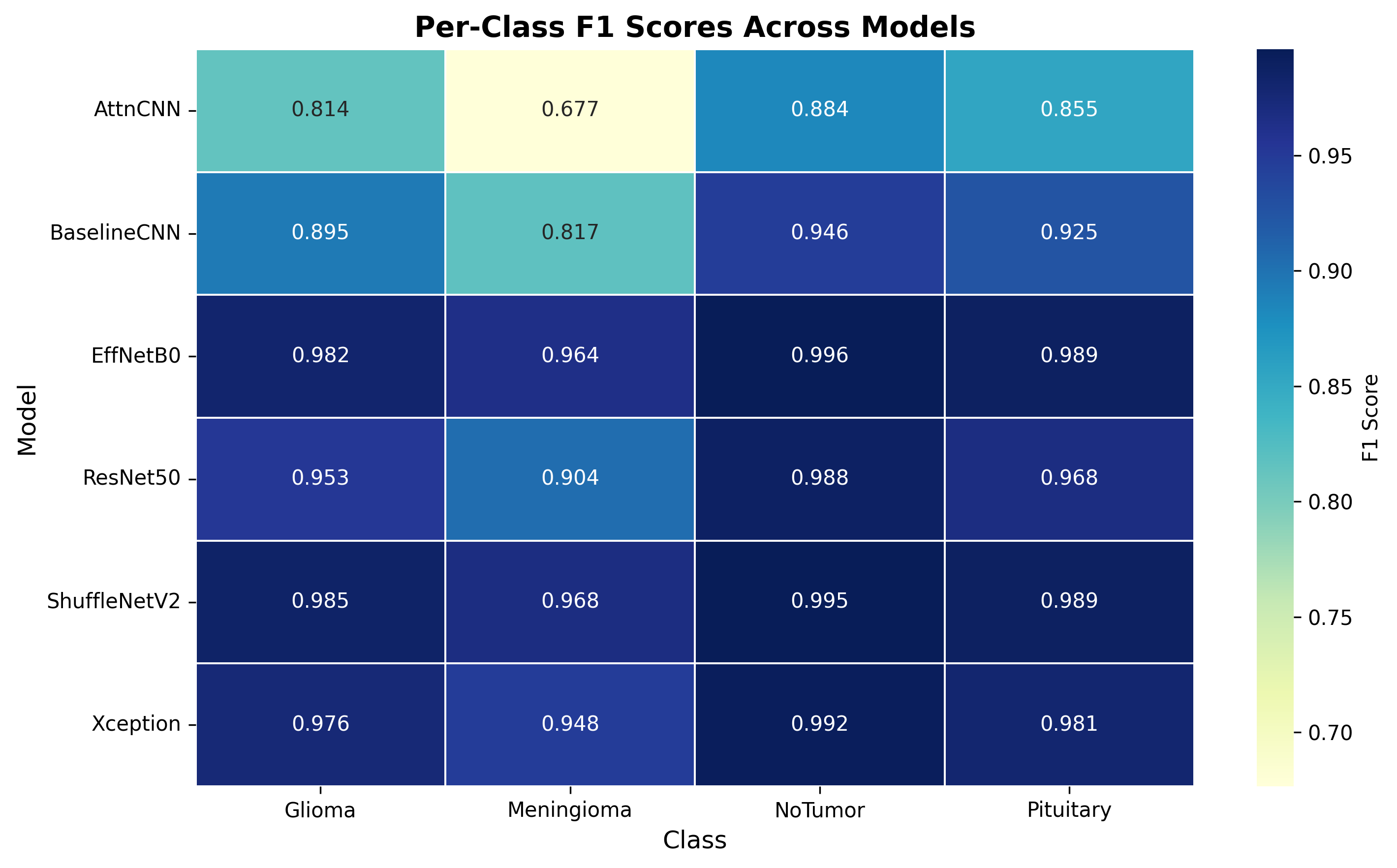}
    \caption{Per-class F1-scores of all models under 5-fold group cross-validation.}
    \label{fig:kfold-f1}
\end{figure}

Per-class F1-scores (Fig.~\ref{fig:kfold-f1}) show that EfficientNet-B0 and ShuffleNetV2 achieved consistently high performance across all classes, exceeding 0.98 for the glioma and no-tumor classes and 0.96 for meningioma and pituitary tumors. In contrast, the attention CNN degraded markedly, particularly for meningioma (F1 = 0.677).

\begin{figure}[!htbp]
    \centering
    \includegraphics[width=0.85\linewidth]{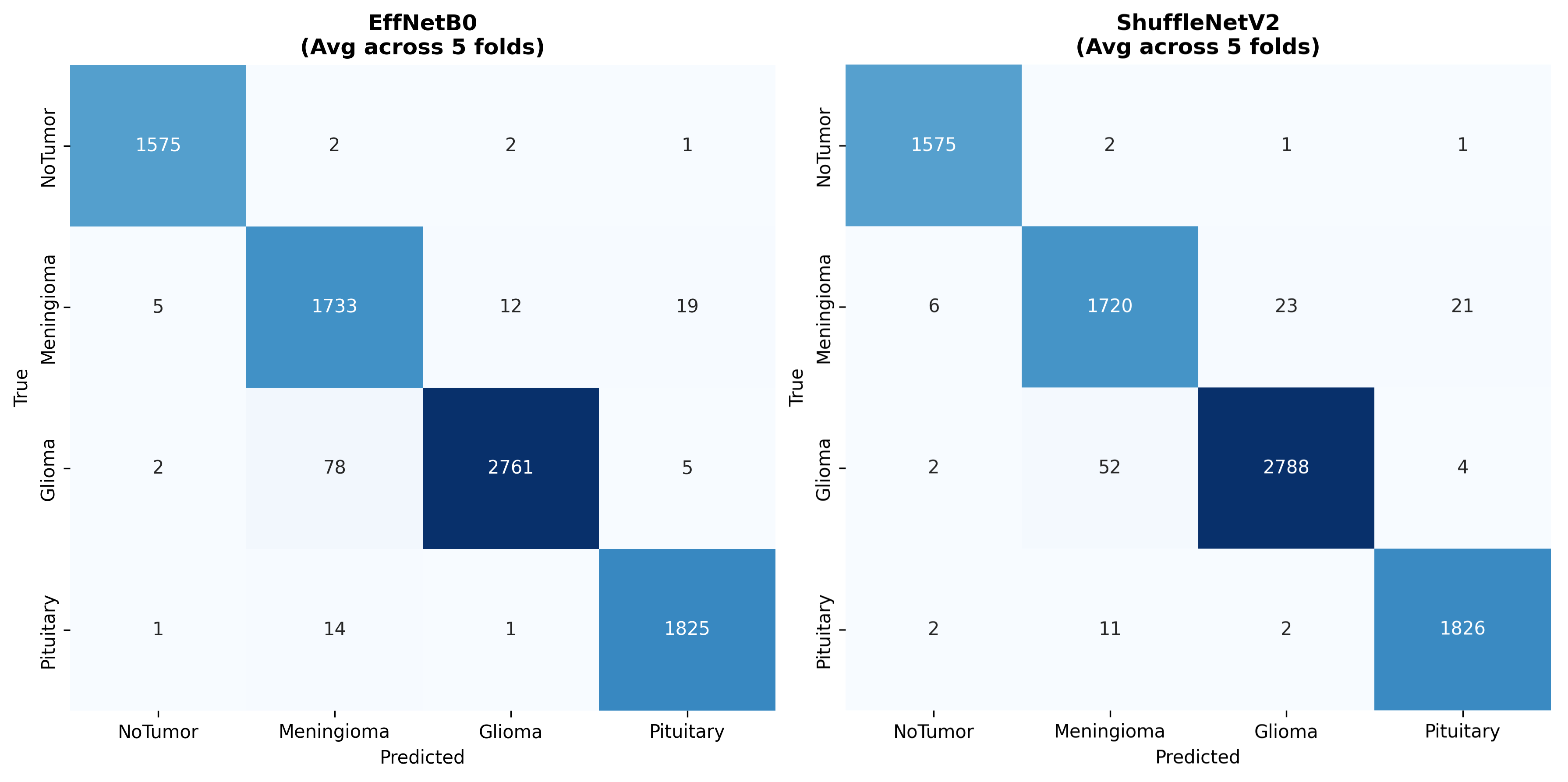}
    \caption{Confusion matrices averaged across the five folds: (a) EfficientNet-B0, (b) ShuffleNetV2.}
    \label{fig:kfold-cm}
\end{figure}

The confusion matrices of the two best models (Fig.~\ref{fig:kfold-cm}) show minimal inter-class confusion, particularly between glioma and pituitary tumors, confirming strong class separability.

\subsection{Cross-Dataset Generalization (LODO Evaluation)}
To evaluate cross-dataset generalization, the LODO protocol (Section~\ref{sec:lodo}) was applied: the Lakshan dataset was held out entirely for testing, and all remaining datasets were used for training. LODO provides a stringent estimate of real-world performance, in which test data come from a source never seen during training.

\begin{table}[!htbp]
\centering
\caption{Cross-dataset generalization under the leave-one-dataset-out (LODO) protocol.}
\label{tab:lodo}
\begin{tabular}{lcc}
\toprule
\textbf{Model} & \textbf{LODO accuracy} & \textbf{LODO F1-score} \\
\midrule
EfficientNet-B0 & \textbf{0.784} & \textbf{0.782} \\
Xception        & 0.766 & 0.764 \\
ResNet-50       & 0.717 & 0.709 \\
ShuffleNetV2    & 0.662 & 0.635 \\
Baseline CNN    & 0.653 & 0.636 \\
Attention CNN   & 0.565 & 0.542 \\
\bottomrule
\end{tabular}
\end{table}

\begin{figure}[!htbp]
    \centering
    \includegraphics[width=0.9\linewidth]{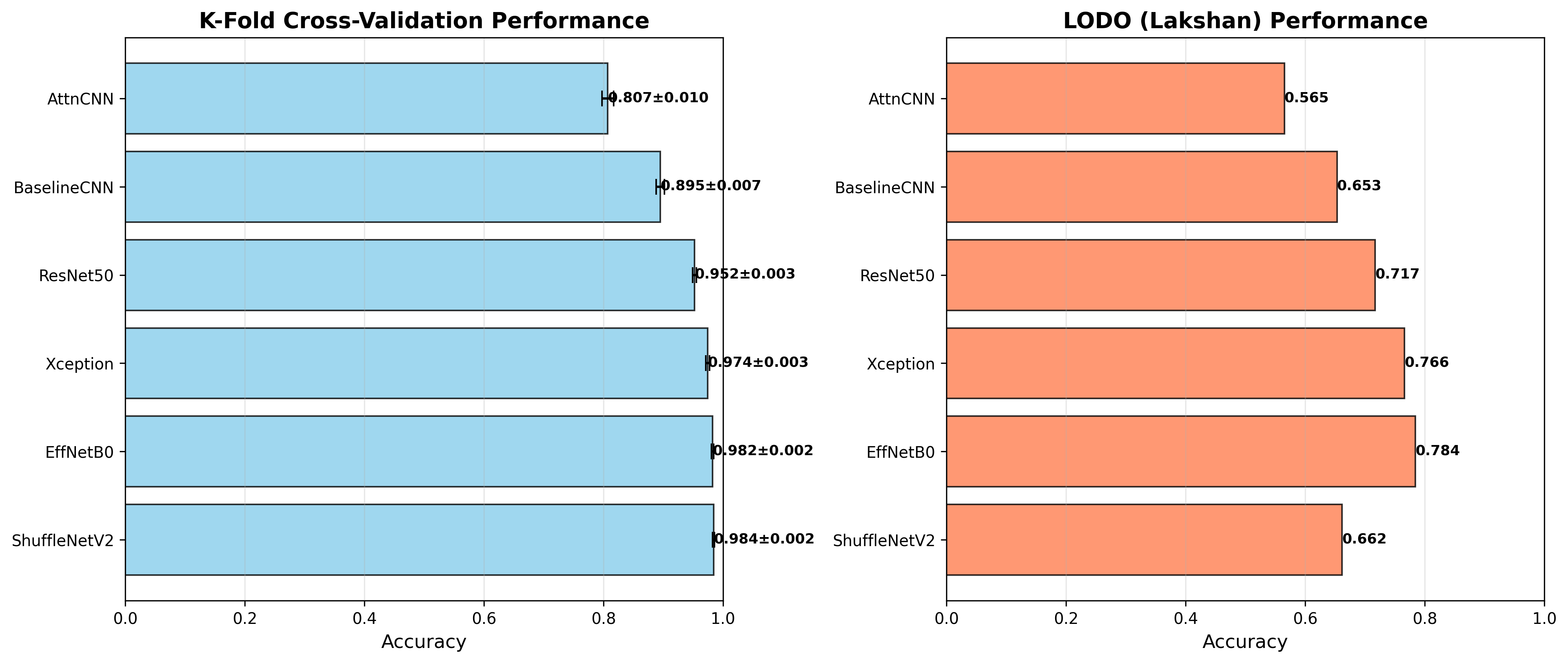}
    \caption{Accuracy of all models under (a) 5-fold group cross-validation and (b) LODO evaluation with the Lakshan dataset held out.}
    \label{fig:model-comparison}
\end{figure}

Table~\ref{tab:lodo} and Fig.~\ref{fig:model-comparison} present the LODO results. EfficientNet-B0 achieved the highest LODO accuracy (0.784) and F1-score (0.782), followed by Xception (0.766 and 0.764, respectively). In contrast, the lightweight and custom models degraded substantially, with the attention CNN dropping to 0.565 accuracy.

Compared with K-fold validation, every model lost performance, confirming the presence of domain shift between data sources. EfficientNet-B0 and Xception were markedly more robust than the other models, indicating a greater ability to generalize to unseen data sources and imaging protocols.

\subsection{Statistical Significance and Model Comparison}
Paired t-tests on the per-fold K-fold accuracies were used to assess whether the observed differences were statistically significant (Table~\ref{tab:stat-tests}). Every tested comparison between EfficientNet-B0, ShuffleNetV2 or Xception and the custom CNNs was significant ($p < 10^{-4}$), as was the difference between ShuffleNetV2 and ResNet-50. The difference between EfficientNet-B0 and ShuffleNetV2 was not significant ($p = 0.19$), indicating statistically equivalent in-distribution performance. Their LODO results, however, differ substantially, highlighting the stronger cross-dataset generalization of EfficientNet-B0.

\begin{table}[!htbp]
\centering
\caption{Pairwise statistical significance testing (paired t-test on per-fold K-fold accuracy).}
\label{tab:stat-tests}
\begin{tabular}{llcccc}
\toprule
\textbf{Model 1} & \textbf{Model 2} & \textbf{Mean diff.} & \textbf{t-statistic} & \textbf{p-value} & \textbf{Significant ($\alpha = 0.05$)} \\
\midrule
EfficientNet-B0 & Attention CNN & 0.1752 & 43.38 & $1.7\times10^{-6}$ & Yes \\
ShuffleNetV2    & Attention CNN & 0.1770 & 36.68 & $3.3\times10^{-6}$ & Yes \\
Xception        & Attention CNN & 0.1670 & 35.04 & $4.0\times10^{-6}$ & Yes \\
EfficientNet-B0 & Baseline CNN  & 0.0873 & 31.79 & $5.8\times10^{-6}$ & Yes \\
ShuffleNetV2    & Baseline CNN  & 0.0891 & 27.92 & $9.8\times10^{-6}$ & Yes \\
ResNet-50       & ShuffleNetV2  & $-0.0320$ & $-21.42$ & $2.8\times10^{-5}$ & Yes \\
EfficientNet-B0 & ShuffleNetV2  & $-0.0018$ & $-1.57$ & 0.19 & No \\
\bottomrule
\end{tabular}
\end{table}

Table~\ref{tab:model-comparison} compares all models under both the in-distribution (K-fold) and cross-dataset (LODO) settings. ShuffleNetV2 achieved the highest in-distribution performance (98.41\% accuracy, F1-score 0.984), indicating the closest fit to the training distribution. EfficientNet-B0, however, showed the strongest cross-dataset generalization, with a LODO accuracy of 78.4\%, outperforming all other models on the unseen dataset. This reveals a trade-off between peak in-distribution accuracy and robustness to domain shift.

\begin{table}[!htbp]
\centering
\caption{Overall model comparison under K-fold cross-validation and LODO evaluation.}
\label{tab:model-comparison}
\begin{tabular}{lccccc}
\toprule
\textbf{Model} & \textbf{K-fold acc.} & \textbf{K-fold F1} & \textbf{Std acc.} & \textbf{LODO acc.} & \textbf{LODO F1} \\
\midrule
ShuffleNetV2    & \textbf{0.984} & \textbf{0.984} & 0.0016 & 0.662 & 0.635 \\
EfficientNet-B0 & 0.982 & 0.982 & 0.0021 & \textbf{0.784} & \textbf{0.782} \\
Xception        & 0.974 & 0.974 & 0.0031 & 0.766 & 0.764 \\
ResNet-50       & 0.952 & 0.952 & 0.0033 & 0.717 & 0.709 \\
Baseline CNN    & 0.895 & 0.895 & 0.0070 & 0.653 & 0.636 \\
Attention CNN   & 0.807 & 0.807 & 0.0099 & 0.565 & 0.542 \\
\bottomrule
\end{tabular}
\end{table}

\subsection{Per-Class Diagnostic Performance}
Table~\ref{tab:perclass-f1} reports the class-wise F1-scores of the three best models. All three achieve near-perfect performance for the no-tumor and pituitary classes, reflecting their visually distinct anatomical and radiological patterns. Glioma classification is also strong, with F1-scores above 0.975 for all three models. Meningioma remains the most challenging class, with consistently lower F1-scores across architectures. This is clinically expected, as meningiomas can share imaging characteristics with gliomas and pituitary tumors.

\begin{table}[!htbp]
\centering
\caption{Per-class F1-scores of the best-performing models (K-fold evaluation).}
\label{tab:perclass-f1}
\begin{tabular}{lcccc}
\toprule
\textbf{Model} & \textbf{No tumor} & \textbf{Meningioma} & \textbf{Glioma} & \textbf{Pituitary} \\
\midrule
ShuffleNetV2    & 0.995 & \textbf{0.968} & \textbf{0.985} & \textbf{0.989} \\
EfficientNet-B0 & \textbf{0.996} & 0.964 & 0.982 & \textbf{0.989} \\
Xception        & 0.992 & 0.948 & 0.976 & 0.981 \\
\bottomrule
\end{tabular}
\end{table}

\subsection{Error Analysis and Failure Modes}
To understand the limitations of the best in-distribution model, a detailed error analysis of ShuffleNetV2 was performed using its pooled out-of-fold predictions, absolute and normalized confusion matrices, and misclassification statistics. The analysis reveals systematic patterns in its failures.

\subsubsection{Confusion Matrix Interpretation}
\begin{figure}[!htbp]
    \centering
    \includegraphics[width=\linewidth]{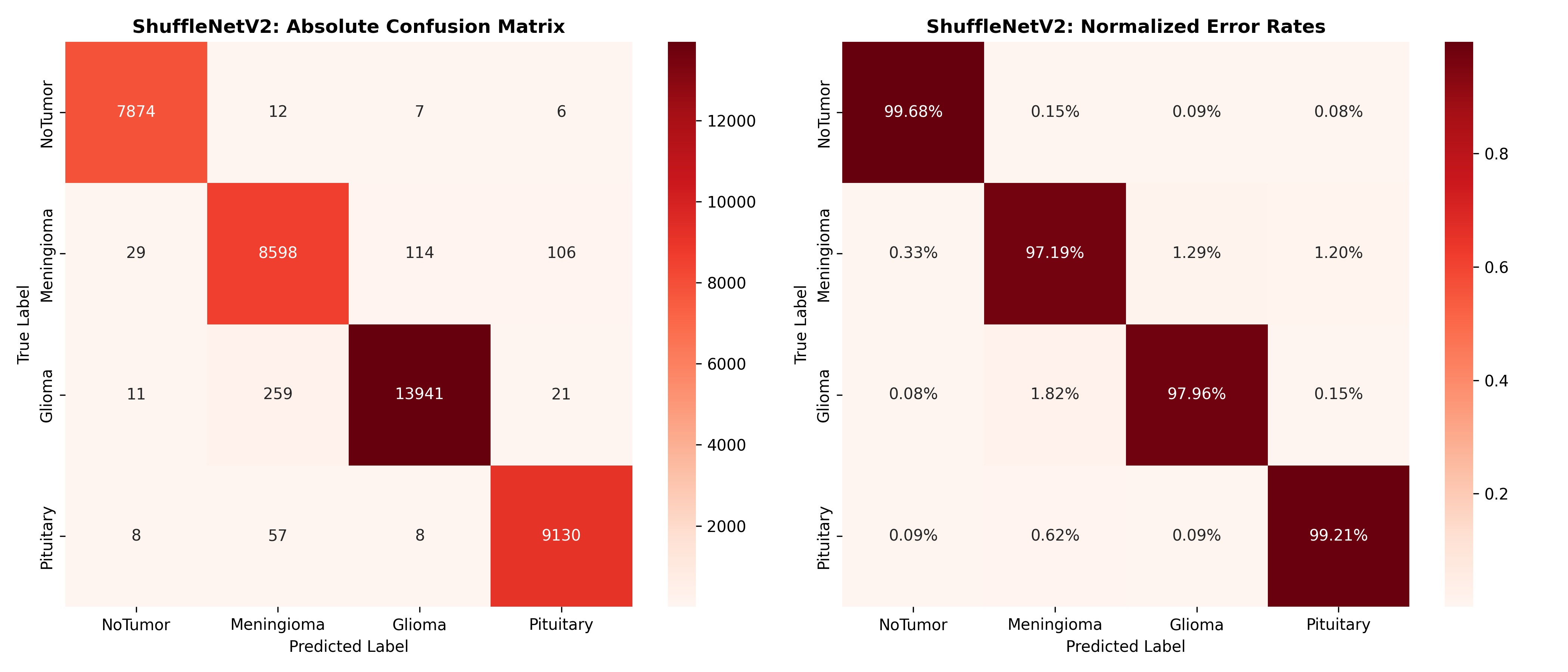}
    \caption{Error analysis of ShuffleNetV2: absolute confusion matrix (left) and row-normalized rates (right), pooled over the five folds.}
    \label{fig:error-shufflenet}
\end{figure}

Fig.~\ref{fig:error-shufflenet} presents the absolute and normalized confusion matrices of ShuffleNetV2. The model achieved near-perfect discrimination for no tumor (99.68\%) and pituitary tumors (99.21\%), indicating that healthy brains and pituitary tumors have highly distinctive visual patterns that the network captures robustly.

Glioma is also classified with very high accuracy (97.96\%), suggesting that tumors with strong contrast and irregular margins are reliably detected. Meningioma, however, has a comparatively lower recall (97.19\%), with a non-negligible fraction of cases confused with glioma and pituitary tumors.

This asymmetry is clinically meaningful: meningiomas often present imaging features that partially overlap with gliomas (tumor mass and edema) and with pituitary tumors (location and shape), making them intrinsically harder to distinguish, even for radiologists.

\subsubsection{Dominant Error Modes}
\begin{figure}[!htbp]
    \centering
    \includegraphics[width=0.85\linewidth]{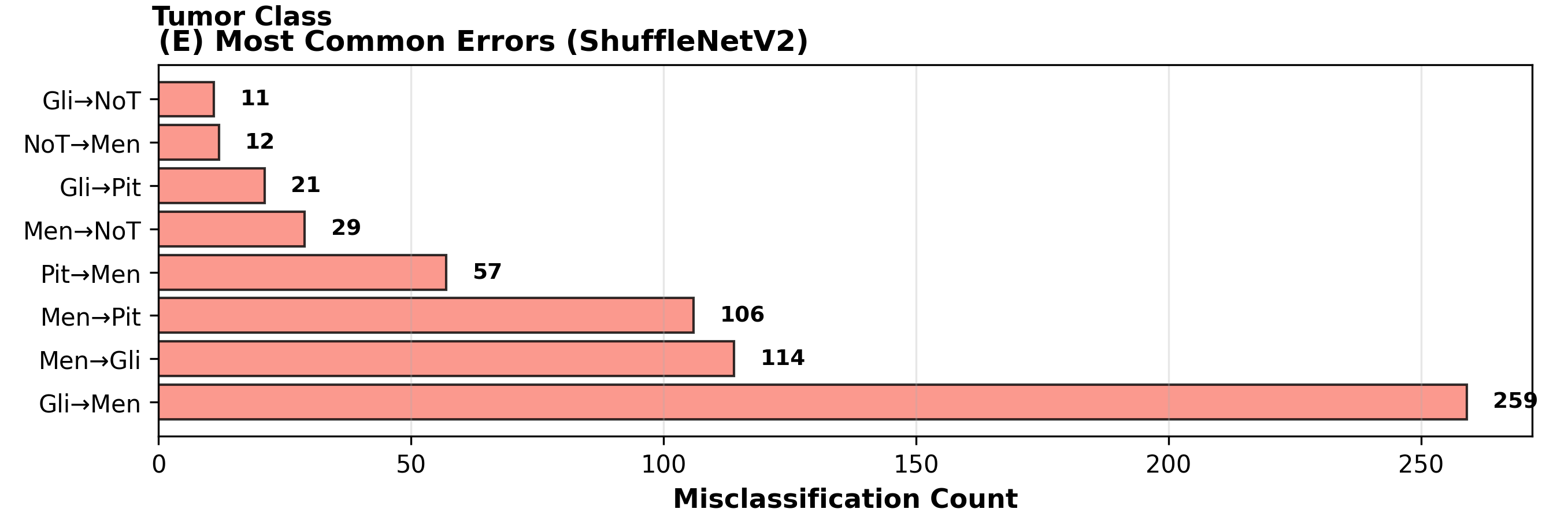}
    \caption{Most frequent misclassification pairs of ShuffleNetV2.}
    \label{fig:top-errors}
\end{figure}

Table~\ref{tab:error-pairs} and Fig.~\ref{fig:top-errors} summarize the most frequent misclassification pairs.
They reveal strong bidirectional confusion between glioma and meningioma, which accounts for the largest share of errors. Although ShuffleNetV2 is highly accurate overall, it struggles when tumors have ambiguous boundaries, similar texture or otherwise overlapping characteristics.

\begin{table}[!htbp]
\centering
\caption{Misclassification patterns of ShuffleNetV2 (error rate normalized by the number of true-class samples).}
\label{tab:error-pairs}
\begin{tabular}{llcc}
\toprule
\textbf{True class} & \textbf{Predicted class} & \textbf{Error count} & \textbf{Error rate} \\
\midrule
Glioma      & Meningioma & 259 & 0.0182 \\
Meningioma  & Glioma     & 114 & 0.0129 \\
Meningioma  & Pituitary  & 106 & 0.0120 \\
Pituitary   & Meningioma & 57  & 0.0062 \\
Meningioma  & No tumor   & 29  & 0.0033 \\
Glioma      & Pituitary  & 21  & 0.0015 \\
No tumor    & Meningioma & 12  & 0.0015 \\
Glioma      & No tumor   & 11  & 0.0008 \\
Pituitary   & No tumor   & 8   & 0.0009 \\
Pituitary   & Glioma     & 8   & 0.0009 \\
No tumor    & Glioma     & 7   & 0.0009 \\
No tumor    & Pituitary  & 6   & 0.0008 \\
\bottomrule
\end{tabular}
\end{table}

The meningioma--pituitary confusions are likely driven by tumor proximity and similar contrast-enhancement patterns, especially in low-resolution or motion-affected scans. Importantly, the no-tumor class had very low confusion rates with every tumor type ($<0.2\%$), supporting the model's reliability in ruling out disease.

\subsection{Effect of Architectural Choices}
To assess the contribution of different architectural choices, the six models were compared as progressively more sophisticated designs: the baseline CNN, the attention CNN, ResNet-50, Xception, EfficientNet-B0 and ShuffleNetV2.

The baseline CNN serves as the control model, with no pre-trained weights or attention. The attention CNN adds an attention module to test whether explicit spatial focus improves tumor discrimination. The remaining four models are established backbone architectures; EfficientNet-B0, ResNet-50 and Xception are initialized from ImageNet weights, which allows the reuse of features learned on large natural-image datasets.

Under leakage-aware evaluation, adding attention to a model trained from scratch did not help: the attention CNN scored lower than the baseline CNN in both K-fold (0.807 versus 0.895) and LODO (0.565 versus 0.653) evaluation. Attention alone was therefore insufficient, and possibly detrimental, when trained from scratch on this data. In contrast, all four backbone architectures outperformed both custom CNNs in K-fold evaluation (significantly so in every pair tested; Table~\ref{tab:stat-tests}), and the three ImageNet-pretrained models also generalized best under LODO.

Model size was not the deciding factor. ResNet-50, the largest backbone, trailed the lightweight ShuffleNetV2 in K-fold accuracy (0.952 versus 0.984), whereas ShuffleNetV2 trailed ResNet-50 under LODO (0.662 versus 0.717). EfficientNet-B0 offered the best overall balance: it was statistically tied with ShuffleNetV2 in-distribution and achieved the highest cross-dataset accuracy, suggesting that compound scaling combined with pre-trained representations gives the best trade-off between efficiency and robustness.

Overall, the choice of backbone and pre-training accounted for most of the performance differences, whereas adding attention to a from-scratch CNN provided no benefit in these experiments.

\subsection{Explainability via Grad-CAM}
\begin{figure}[!htbp]
    \centering
    \begin{subfigure}[b]{0.85\linewidth}
        \includegraphics[width=\linewidth]{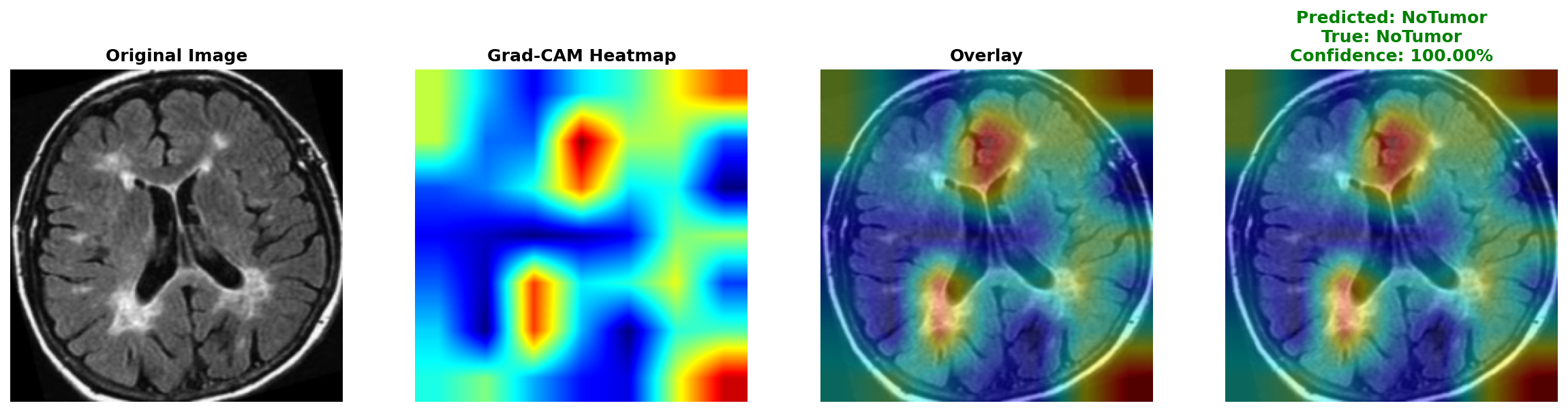}
        \caption{}
    \end{subfigure}

    \begin{subfigure}[b]{0.85\linewidth}
        \includegraphics[width=\linewidth]{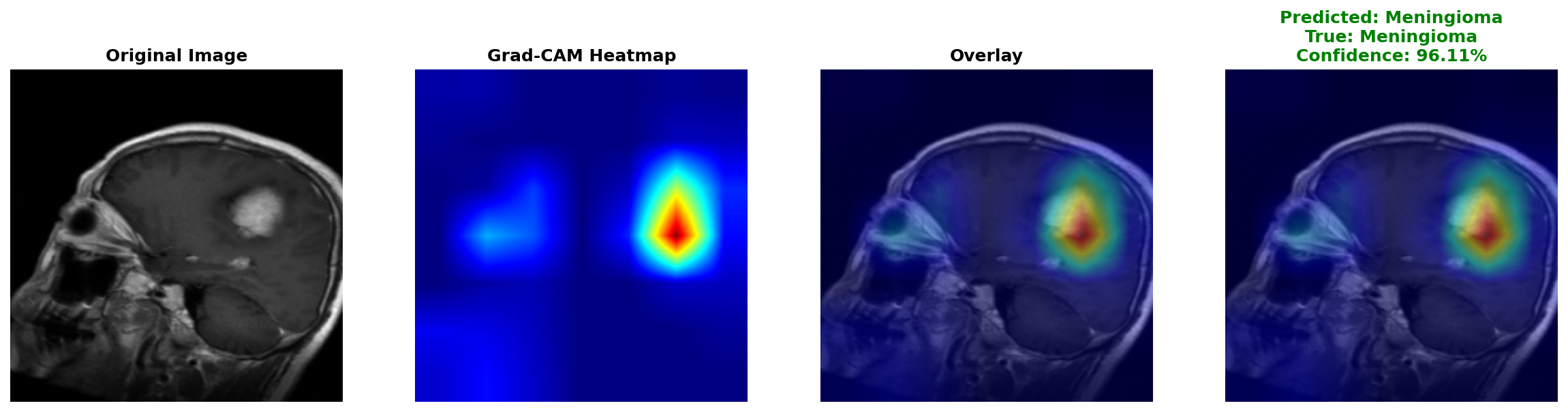}
        \caption{}
    \end{subfigure}

    \begin{subfigure}[b]{0.85\linewidth}
        \includegraphics[width=\linewidth]{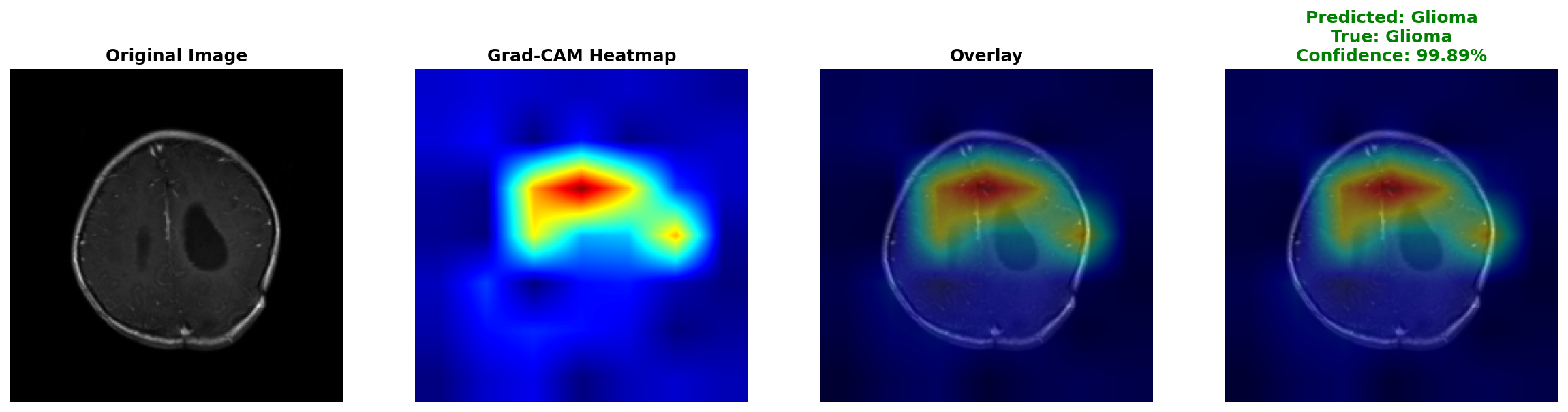}
        \caption{}
    \end{subfigure}

    \begin{subfigure}[b]{0.85\linewidth}
        \includegraphics[width=\linewidth]{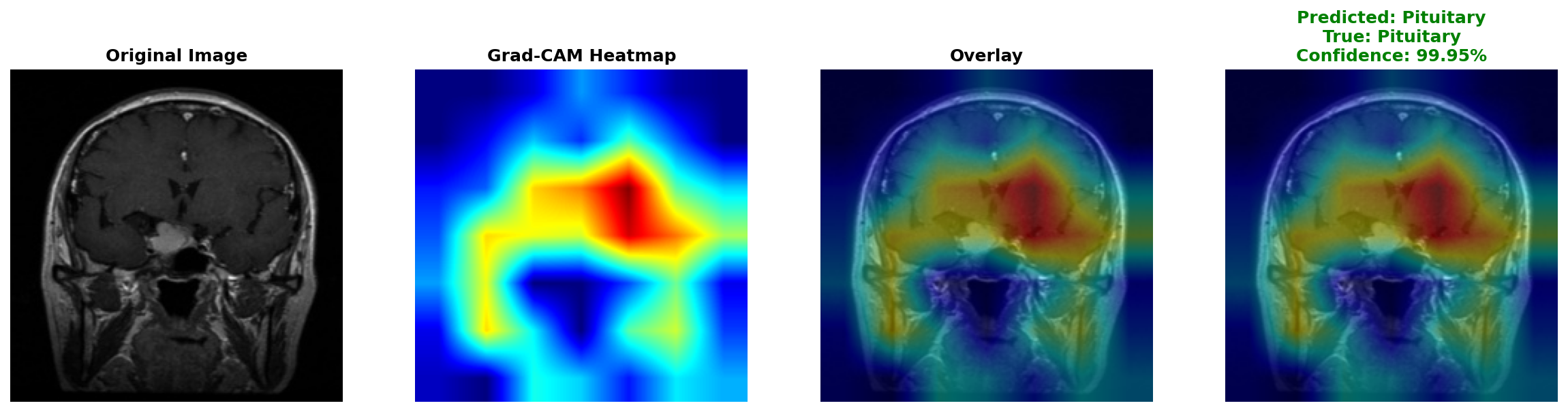}
        \caption{}
    \end{subfigure}
    \caption{Grad-CAM visualizations for ShuffleNetV2: (a) no tumor, (b) meningioma, (c) glioma, (d) pituitary tumor.}
    \label{fig:gradcam-shufflenet}
\end{figure}

To interpret the decision-making process and check its clinical plausibility, Grad-CAM was applied to the best in-distribution model, ShuffleNetV2. Grad-CAM produces heatmaps of the spatial regions that contribute most to a prediction, allowing verification that the model attends to anatomically meaningful tumor regions.
Representative examples from all four classes (no tumor, meningioma, glioma and pituitary tumor) are shown in Fig.~\ref{fig:gradcam-shufflenet}.

For no-tumor cases, the Grad-CAM maps show diffuse, low-intensity activation without strong focal hotspots, indicating that the model recognizes the absence of pathological structures rather than relying on spurious artifacts.

For meningioma samples, activation concentrates strongly on the well-defined extra-axial tumor mass near the skull boundary, consistent with the radiological appearance of meningiomas. This suggests that the model learns shape- and boundary-driven features rather than global texture patterns.

For glioma, Grad-CAM highlights infiltrative intra-axial regions with irregular margins, closely matching the heterogeneous and invasive morphology of gliomas. Because glioma and meningioma can share intensity profiles despite their different spatial growth patterns, this is also consistent with the higher glioma--meningioma confusion observed in the error analysis.

For pituitary tumors, the heatmaps are sharply localized around the sellar and suprasellar regions, corresponding to the anatomical location of pituitary adenomas. These compact, centered activations are consistent with the high classification accuracy for this class.

Overall, the Grad-CAM visualizations indicate that ShuffleNetV2 bases its predictions on clinically relevant tumor regions rather than on background or skull artifacts. This supports the view that its high K-fold performance reflects meaningful anatomical learning. Its weaker LODO performance, however, shows that attending to the right regions does not by itself guarantee robustness to a new data source.

\subsection{Discussion}
Under the standard split, the baseline CNN was generally confident ($>0.9$) in its correct classifications but produced some low-confidence misclassifications, particularly between tumor subtypes, which motivated the stronger feature extractors evaluated subsequently. Grad-CAM maps of the attention CNN concentrated on regions corresponding to visible tumor masses, providing spatial interpretability, and its correct predictions received very high confidence scores (0.999). ShuffleNet was likewise highly confident in most correct predictions ($>0.9$), while assigning lower confidence (0.252--0.292) to ambiguous cases, and its probability distributions showed clear class separation.

The leakage-aware experiments, however, change the overall picture. Once exact and near-duplicate images were removed and similar images were kept within folds, the pretrained backbones and ShuffleNetV2 clearly outperformed the custom CNNs, and the attention module no longer improved on the baseline. When an entire data source was held out, every model lost between roughly 20 and 32 percentage points of accuracy, and the in-distribution ranking did not carry over: ShuffleNetV2, the best K-fold model, ranked fourth under LODO, while EfficientNet-B0 generalized best. Accuracy measured on random or K-fold splits of the training sources should therefore not be taken as an estimate of clinical performance.

Several limitations should be considered when translating these results to clinical practice:
\begin{itemize}
    \item \textbf{Data provenance:} all four datasets are publicly available collections without patient identifiers or scanner metadata, so patient-level separation could only be approximated through image hashing, and the datasets may not capture the full range of imaging protocols, scanners and populations encountered clinically.
    \item \textbf{Limited cross-dataset testing:} only one dataset could serve as a held-out test source, because it is the only one containing all four classes; generalization to other sources remains to be confirmed.
    \item \textbf{Class balance:} despite mitigation strategies, residual class imbalance may introduce biases that affect real-world performance where tumor prevalence differs from the training data.
    \item \textbf{Single-modality, 2D input:} clinical diagnosis typically incorporates patient history, symptoms, additional MRI sequences and volumetric information, none of which were included in this slice-based evaluation.
\end{itemize}

\section{Conclusion and Future Work}
This study compared six deep learning architectures for four-class brain tumor classification on a corpus assembled from four public MRI datasets, under three evaluation protocols. Under a conventional random split, the efficient ShuffleNet achieved the highest accuracy (97.1\%) with only 1.2M parameters, and the models trained from scratch appeared to outperform transfer learning. Under leakage-aware group K-fold cross-validation, this ranking reversed: ShuffleNetV2 (98.41\%) and EfficientNet-B0 (98.23\%) were statistically indistinguishable and both significantly outperformed the custom CNNs, and the attention module did not improve on the baseline CNN. When an entire dataset was held out, all models degraded substantially; EfficientNet-B0 generalized best (78.4\%), whereas ShuffleNetV2 fell to 66.2\%.

These findings show that the choice of evaluation protocol can change both the absolute performance and the ranking of models, and that deduplication, grouped splitting and cross-dataset testing are necessary for realistic performance estimates. Among the models evaluated, EfficientNet-B0 offers the best combination of in-distribution accuracy, cross-dataset robustness and computational efficiency, while ShuffleNetV2 remains attractive where computational resources are very limited and the deployment data closely match the training data. The persistent glioma--meningioma confusion across all architectures reflects the inherent difficulty of this distinction and positions AI as a complementary tool rather than a replacement for clinical expertise.

Building on the current 2D framework, future work will develop 3D volumetric analysis for glioma subtype classification coupled with cognitive impact assessment, addressing several clinical needs:
\begin{itemize}
    \item \textbf{3D volumetric architecture:} extending efficient architectures to 3D convolutions will capture volumetric tumor characteristics, growth patterns and spatial relationships with eloquent brain regions. A 3D approach enables analysis of tumor heterogeneity, necrotic regions and enhancement patterns that are crucial for glioma grading (WHO grades 1--4) and molecular subtyping (IDH-mutant versus IDH-wildtype, 1p/19q codeletion status).
    \item \textbf{Cognitive function correlation:} integrating neuropsychological assessment data with 3D imaging will enable prediction of cognitive deficits from tumor location, size and proximity to functional areas. Such a multimodal approach could identify patients at risk of language, memory or executive-function impairment, guiding treatment decisions and rehabilitation planning.
    \item \textbf{Generalization and clinical implementation:} the 3D system will retain the computational efficiency principles established here and will be validated on multiple independent, held-out data sources to ensure deployment feasibility in clinical settings.
\end{itemize}

\bibliographystyle{cas-model2-names}
\bibliography{cas-refs}

\end{document}